\documentclass[preprint,10pt,a4paper]{elsarticle}
\PassOptionsToPackage{no-math}{fontspec}
\usepackage{amsmath,amssymb}
\usepackage{fontspec}
\usepackage{unicode-math}
\usepackage{graphicx,xcolor}
\usepackage{booktabs,array,tabularx,makecell}
\usepackage{enumitem}
\usepackage{geometry}
\usepackage{setspace}
\usepackage[section]{placeins}
\usepackage{caption}
\usepackage{etoolbox}
\biboptions{numbers,sort&compress}
\usepackage{xurl}
\usepackage[unicode,hidelinks]{hyperref}
\hypersetup{pdftitle={Visual Sufficiency Landscapes: Understanding Query-Conditioned Reliability in Visual Token Pruning},pdfsubject={arXiv preprint}}

\makeatletter
\patchcmd{\pprintMaketitle}{\Large\@title}{\fontsize{14}{17}\selectfont\@title}{}{}
\patchcmd{\pprintMaketitle}{\normalsize\elsauthors}{\footnotesize\elsauthors}{}{}
\patchcmd{\@author}{\normalsize}{\footnotesize}{}{}
\patchcmd{\pprintMaketitle}{\elsaddress\par\vskip36pt}{\elsaddress\par\vskip12pt}{}{}
\patchcmd{\pprintMaketitle}{\thispagestyle{pprintTitle}}{\thispagestyle{plain}}{}{}
\makeatother
\AtBeginEnvironment{table}{\normalsize}
\newcommand{\psuf}{p_{\mathrm{suf}}}
\newtheorem{proposition}{Proposition}

\date{}

\begin{document}
\begin{frontmatter}
\title{Visual Sufficiency Landscapes: Understanding Query-Conditioned Reliability in Visual Token Pruning}
\author{Zhenghan Li}
\author{Kexin Sun}
\author{Wei Zheng\corref{corresponding}}
\ead{wzheng@nwpu.edu.cn}
\cortext[corresponding]{Corresponding author.}
\author{Peng Hu}
\author{Xiao Wei}
\author{Anbo Huang}
\affiliation{organization={School of Software, Northwestern Polytechnical University},
  citysep={}, city={Xi'an}, postcode={710129}, country={China}}
\begin{abstract}
\singlespacing
Different visual token pruning strategies retain different subsets while maintaining similar task performance. This motivates a combinatorial question: how many visual subsets can preserve model behavior, and how does their abundance depend on the query? We introduce \emph{visual sufficiency landscapes} to characterize this abundance. At a fixed layer and budget, visual sufficiency probability is the fraction of uniformly sampled subsets that jointly preserve the dense model's answer, reference-answer likelihood, and output distribution. Independent sampling replicates separate input variation from estimation noise. Across Qwen2.5-VL-7B, InternVL2-8B, and LLaVA-v1.5-7B, differences between inputs on GQA and TextVQA exceed replication error, and different questions about the same GQA image produce distinct landscapes. Sufficiency probability also improves failure prediction for full pruning methods. On LLaVA, adding this probability increases AUC by 0.1151 for FastV, 0.0683 for VisionZip, and 0.1401 for SparseVLM, while reducing log loss and Brier score. Several architecture-adapted configurations on the other two models reproduce these gains. Under controlled selection rules, predictors trained on the original images reduce log loss on new images by 9.0\%, 13.0\%, and 14.1\%, respectively. These results establish successful-subset abundance as a measurable dimension of local prunability and connect it to the identification of high-risk pruning conditions.
\end{abstract}
\end{frontmatter}
\thispagestyle{plain}

\singlespacing
\section{Introduction}
\label{sec:introduction}
Vision-language models connect image content to language models through visual tokens~\cite{liu2024llava,bai2025qwen}. Longer visual sequences increase inference cost, making visual token pruning an important route to efficient multimodal inference. Attention-based filtering, redundancy reduction, and diversity-based selection construct retained sets using different criteria, yet all can reduce visual computation while maintaining task performance close to that of the dense model~\cite{chen2024fastv,yang2024visionzip,alvar2025divprune}. Why can different retained visual subsets succeed? This motivates us to examine the multiplicity of effective visual combinations and the flexibility in subset selection for a given input.

A natural explanation is that multiple visual combinations can support the same task. A successful pruning decision may occur where many subsets preserve behavior, or it may depend on a few viable combinations. The selected sets succeed in both cases, but the flexibility in selection differs. This raises a combinatorial question: for a fixed input, layer, and budget, how abundant are successful combinations? Different questions about the same image require different visual information, so how does this abundance change with the query?

Existing research examines input conditions and pruning behavior from several perspectives. Query-conditioned methods incorporate instruction relevance into subset construction~\cite{zhang2025cdpruner,he2026qcpruner}. Information Horizon and CoRePrune examine how visual information changes with depth and how removability depends on the surrounding deletion context, respectively~\cite{wang2026horizon,he2026coreprune}. Research on confidence further connects pruned outputs to risk-controlled budgets~\cite{tan2026confidence}. We study the candidate subset space to characterize how many retention choices succeed under a given condition, providing a combinatorial reference for understanding when different selection strategies can work.

\begin{figure}[!htbp]
\centering
\includegraphics[width=\linewidth]{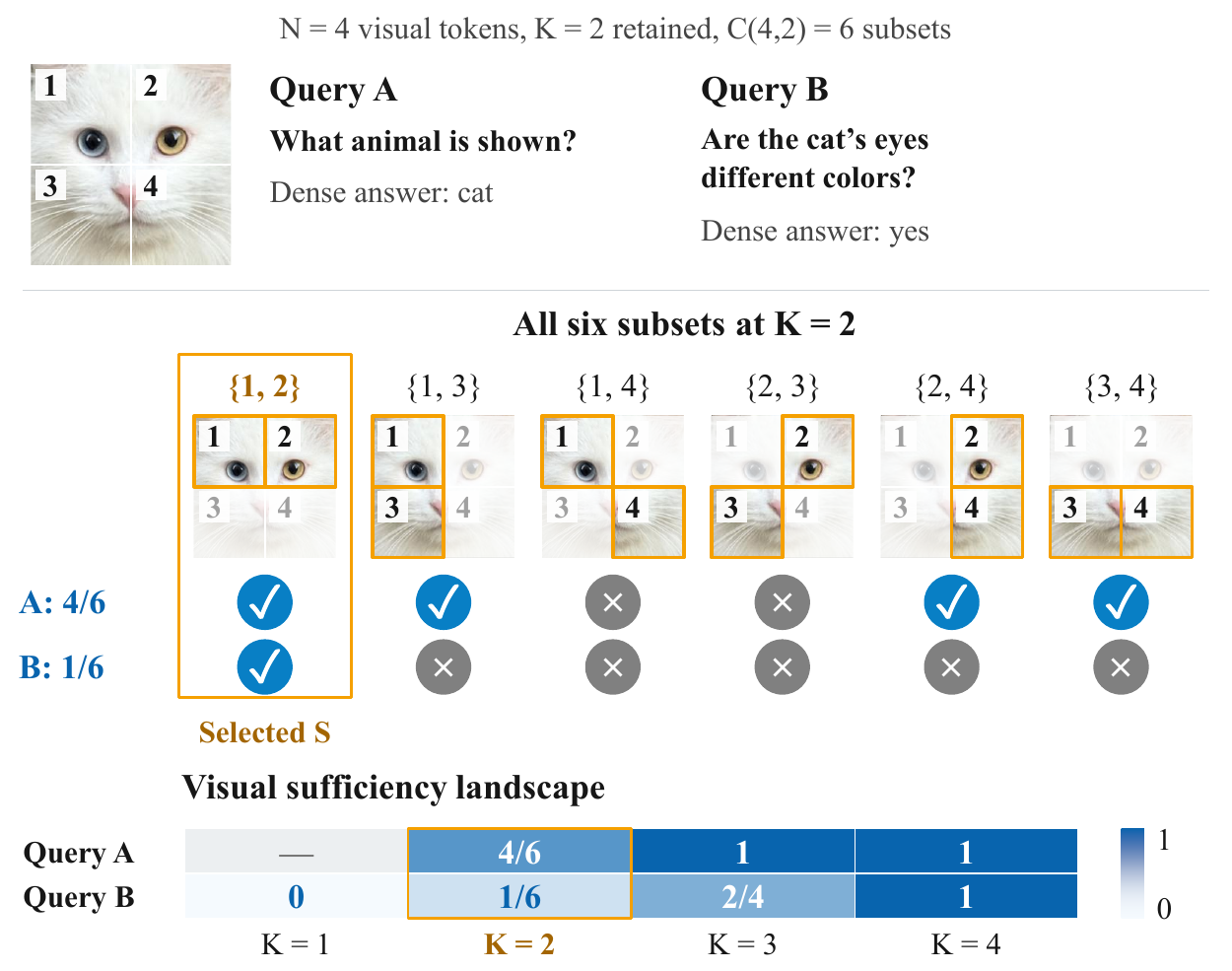}
\caption{The same subset can succeed under different sufficiency probabilities. For a fixed image and layer, $N=4$ and $K=2$ yield six subsets. The orange-framed subset $S=\{1,2\}$ satisfies joint preservation for both questions, yet the fractions of successful subsets are $4/6$ and $1/6$. The heatmap shows a query--budget slice at a fixed layer; gray cells are unspecified. Image patches, success labels, and probabilities are conceptual illustrations.}
\label{fig:teaser}
\end{figure}

We introduce \emph{visual sufficiency landscapes}, which characterize local prunability through the proportion of successful subsets in a fixed-budget combinatorial space. Given an image, question, intervention layer, and retention count, we uniformly sample visual subsets of a fixed size. We define \emph{visual sufficiency probability}, $\psuf$, as the probability of jointly preserving the dense model's answer, reference-answer likelihood, and output distribution. This joint event captures both the generated answer and probabilistic behavior, while the uniform measure provides a common combinatorial reference across inputs. Higher sufficiency probability indicates more abundant successful combinations. Arranging these probabilities across layers and budgets yields a landscape (Fig.~\ref{fig:teaser}), extending the evaluation of individual pruning outcomes to a measurement of successful-subset abundance.

We first use two independent sampling replicates to distinguish input variation from replication error. Multiple questions about each image then hold visual content fixed while varying the query condition. Across three models, between-input curve differences exceed replication error. In the GQA~\cite{hudson2019gqa} multi-question experiment, all 32 images in each model have positive point estimates of the net query curve difference. Supplementary experiments find query differences at all three tested budgets. These results quantify query dependence as variation in the space of successful combinations.

Specific selectors favor particular regions of the subset space. We therefore use paired predictors to test whether sufficiency probability under the uniform measure provides additional risk information. On LLaVA, adding sufficiency probability improves AUC, log loss, and Brier score for predicting joint-preservation failure under FastV, VisionZip, and SparseVLM. On a separate image cohort, alerts based on low sufficiency probability yield an equally weighted risk difference of $0.1719$ across the three methods. Several architecture-adapted configurations on Qwen2.5-VL and InternVL2 reproduce the predictive gains. Experiments with controlled selection rules further show that sufficiency probability adds information beyond single-token information, visual-removal responses, and query-representation summaries. These gains persist when fixed predictors are evaluated on new images, connecting combinatorial measurements to the identification of high-risk inputs for specific methods.

Our contributions are threefold:
\begin{itemize}
\item We quantify the flexibility in choosing effective visual combinations. A joint-preservation event and a uniform subset measure define visual sufficiency landscapes, making successful-combination abundance at a fixed budget an estimable probability. Differences between query-specific probabilities bound the fraction of subsets with discordant preservation outcomes.
\item We quantify query dependence in the space of successful combinations. Independent replicates distinguish input variation from sampling fluctuations, revealing differences across inputs and across questions about the same image in three models and at multiple budgets.
\item We establish the incremental risk information provided by this measure. Full pruning methods, controlled selection rules, replication across architectures, direct evaluation on new images, and cross-image risk alerts empirically connect local prunability to method-specific reliability.
\end{itemize}

Code is publicly available at: {\urlstyle{same}\url{https://github.com/LeeEdward02/visual-sufficiency-landscapes}}.
\section{Related Work}
\label{sec:related}
\subsection{Visual token compression and query-conditioned selection}
Visual token compression reduces multimodal computation by selecting, merging, or reconstructing retained representations. ToMe merges tokens by feature similarity, providing a representative approach to redundancy reduction~\cite{bolya2023tome}. LLaVA-PruMerge and VisionZip apply selection and aggregation to visual representations in multimodal models~\cite{shang2025prumerge,yang2024visionzip}. Within the language model, FastV filters visual positions using attention from early layers~\cite{chen2024fastv}. Subset construction can also follow objectives beyond position-wise attention ranking: DivPrune emphasizes diversity within the retained set~\cite{alvar2025divprune}, while TopV formulates pruning as inference-time optimization and considers compatibility with efficient attention and caching~\cite{yang2025topv}.

The query determines which visual content matters for the current task. PuMer combines text-guided pruning with merging within each modality~\cite{cao2023pumer}. TRIM, SparseVLM, and CROP guide compression using cross-modal relevance, relevant text tokens, and query-relevant regions, respectively~\cite{song2025trim,zhang2025sparsevlm,guo2025crop}. Recent methods incorporate this conditioning into set-level objectives. CDPruner maximizes instruction-conditioned diversity through a determinantal point process~\cite{zhang2025cdpruner}; HAWK combines attention-head importance with text guidance to evaluate visual tokens~\cite{zhu2026hawk}; and QCPruner constructs retained sets through a query-conditioned coverage objective~\cite{he2026qcpruner}.

Budget allocation and subset selection are closely related. ATP-LLaVA adapts retention ratios across instances and layers~\cite{ye2025atp}. FitPrune derives pruning plans for a prescribed budget from attention-distribution statistics~\cite{ye2025fitprune}, and OccamToken uses evidence testing to adapt the retention budget~\cite{li2026occamtoken}. These approaches offer ways to construct pruning actions from input conditions. Our work complements them by measuring the action space itself: at a fixed layer and budget, we estimate the fraction of successful subsets to characterize how many combinations preserve behavior for a given input.

\subsection{Depth and context dependence of visual information}
The propagation of visual information through multimodal computation changes the conditions for compression across layers. VTW exploits the transfer of visual information into text representations to withdraw explicit visual tokens in later layers~\cite{lin2025vtw}. PyramidDrop performs staged compression based on depth-dependent visual redundancy~\cite{xing2025pyramiddrop}. EmbedLens examines input-token sparsity and redundant visual computation through the semantic structure of visual representations~\cite{fan2026embedlens}. These studies extend pruning analysis from input length to representation content and the stage at which it is processed.

More detailed analyses examine how token responses change with context. Information Horizon defines information through the change in target-output probability when retaining a single token relative to removing all visual tokens~\cite{wang2026horizon}. CoRePrune analyzes the effects of representation depth and surrounding deletion sets on removability~\cite{he2026coreprune}. DSTP tracks shifts in relevant visual information during decoding and adapts the retained content accordingly~\cite{kim2026dstp}. We focus on combinatorial responses under fixed intervention conditions and compare them across images and across questions about the same image. Single-position responses and successful-subset abundance are distinct statistical objects: for a general set-response function, observations of the empty set, singletons, and full set do not uniquely determine the success fraction of intermediate-size subsets (Appendix~\ref{app:identifiability}).

\subsection{Subset interventions and predictive sufficiency}
Whether model behavior persists after retaining only part of the input is also central to explanation evaluation. ERASER evaluates sufficiency through prediction changes after retaining a rationale and comprehensiveness through changes after removing it~\cite{deyoung2020eraser}. Recent analyses of rationale learning show that sufficiency scores depend on non-rationale context, and that their relationship to correct classification and learning from rationales varies across tasks and models~\cite{kamp2025rationales}. Sufficiency must therefore be interpreted with respect to the intervention and behavioral reference. We retain a fixed number of ordinary visual positions at an intermediate model layer and jointly evaluate answer preservation, reference-answer likelihood, and output distribution against dense behavior. A uniform subset measure then defines sufficiency probability as the fraction of combinations satisfying joint preservation, enabling comparisons of successful-region abundance across input conditions.

\subsection{Behavioral preservation and pruning risk}
Task performance and behavioral reliability capture different aspects of pruning outcomes. FEATHER shows that preserved performance on some common tasks can coexist with degraded fine-grained abilities such as localization~\cite{endo2025feather}. Research on confidence calibration also distinguishes predictive confidence from empirical accuracy~\cite{guo2017calibration}. For visual pruning, recent work analyzes confident errors following evidence removal and uses post-pruning confidence to trigger full-token inference, with conformal risk control constraining expected deviation from dense outputs~\cite{tan2026confidence}.

These confidence-based approaches assess risk from outputs produced by selected subsets. Visual sufficiency landscapes characterize success probability mass over the candidate subset space at a fixed budget. In Section~\ref{sec:prediction}, we test the incremental predictive information of this combinatorial measure beyond input conditions, dense behavior, and existing information summaries, and examine its value for risk alerts under full pruning methods and on new images. This provides a testable link between successful-combination abundance and the failure risk of specific selection strategies.
\section{Visual Sufficiency Landscapes}
\label{sec:method}
Visual sufficiency landscapes extend the success or failure of an individual pruning decision to the fraction of successful subsets at a fixed budget. This measurement requires a specified subset intervention, a behavioral preservation criterion, and an interpretation of differences under finite sampling. We define the joint-preservation event and sufficiency probability, derive their successful-set interpretation, and use independent replicates as an error reference for landscape comparisons.

\subsection{Visual subset interventions at a fixed budget}
\label{sec:intervention}
\begin{figure}[!htbp]
\centering
\includegraphics[width=\linewidth]{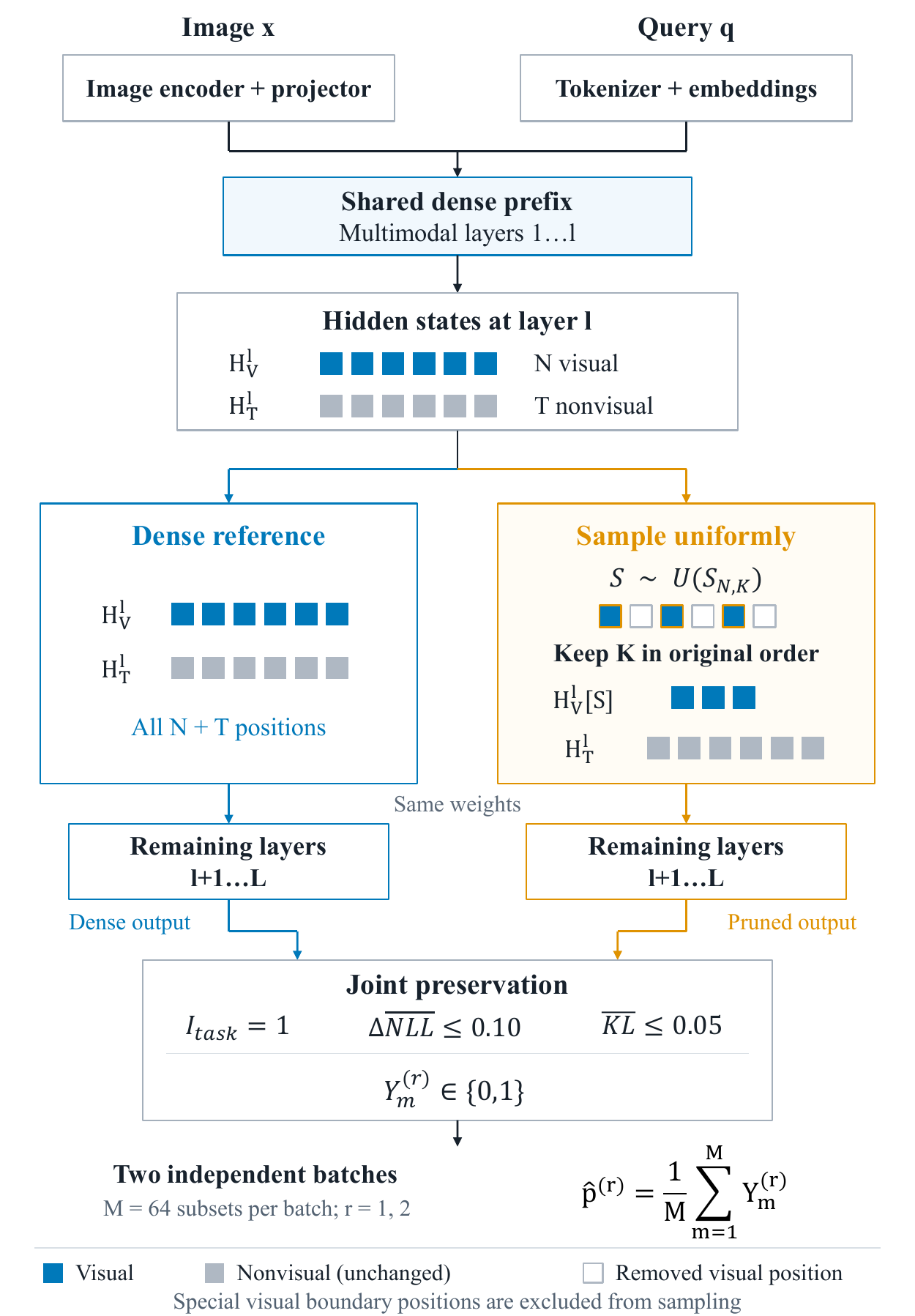}
\caption{Measuring sufficiency probability. A shared dense prefix processes the image and question into layer-$l$ hidden states. Two branches start from these states: the left retains all positions as the dense reference; the right uniformly samples $K$ ordinary visual positions and retains them in their original order. Both preserve the existing nonvisual states and execute subsequent layers with identical parameters. Their outputs are jointly evaluated for answer preservation, mean NLL increase, and mean KL divergence. Two independent batches of $M=64$ interventions estimate $\widehat p^{(r)}$. Special visual boundary positions are excluded from sampling.}
\label{fig:measurement}
\end{figure}

Given an image and question $i=(x,q)$, the visual encoder and connector produce visual representations that enter the language model alongside prompt tokens. Let $H^{l}_{V}$ and $H^{l}_{T}$ denote the visual and nonvisual hidden representations at the specified intervention boundary, and let $N_i$ be the number of ordinary visual tokens eligible for pruning. For retention ratio $\rho$, we use
\begin{equation}
 K_i=\max\{1,\lceil\rho N_i\rceil\},\qquad
 \mathcal S_{N_i,K_i}=\{S\subseteq[N_i]:|S|=K_i\}.
\end{equation}
For $S\in\mathcal S_{N_i,K_i}$, we physically retain the corresponding visual positions, handle positions and masks through the model-specific adaptation, and execute the remaining computation. Figure~\ref{fig:measurement} illustrates this procedure. The prompt, model parameters, and nonvisual positions remain fixed. Special visual boundary tokens are excluded from the sampling pool of ordinary visual positions.

The intervention follows a prefix of multimodal computation, so visual information may already be present in the retained nonvisual hidden representations. Sufficiency therefore characterizes the dependence of subsequent computation on explicit visual positions, conditional on that computational prefix.

\subsection{Sufficiency probability through joint preservation}
\label{sec:sufficiency}
Answer agreement captures visible generation behavior, whereas reference-answer likelihood and teacher-forced output distributions capture probabilistic changes beyond the generated answer. The latter can deviate from the dense model even when pruning leaves the answer unchanged. We therefore combine all three in a \emph{joint-preservation} event. Let $\Delta\overline{\mathrm{NLL}}$ denote the increase in mean answer negative log-likelihood relative to the dense model, and let $\overline{\mathrm{KL}}$ denote the mean KL divergence. A subset succeeds under joint preservation when
\begin{equation}
\begin{split}
Y(i,l,S)=\mathbf 1[&I_{\mathrm{task}}=1,\;
\Delta\overline{\mathrm{NLL}}\leq0.10,\\
&\overline{\mathrm{KL}}\leq0.05].
\end{split}
\label{eq:success}
\end{equation}
$I_{\mathrm{task}}$ requires the normalized generated answer to match the dense answer and the task score to be at least as high as the dense score. Both continuous thresholds are measured in nats per target token, with KL directed from the dense distribution to the pruned distribution. GQA~\cite{hudson2019gqa} uses the normalized reference answer. For TextVQA~\cite{singh2019textvqa}, NLL is weighted by the frequencies of nonempty human answers, and KL uses the most frequent normalized answer. Random interventions with generation or parsing failures count as failures and remain in the denominator.

We define visual sufficiency probability as
\begin{equation}
\psuf(i,l,K_i)=
\mathbb P_{S\sim U(\mathcal S_{N_i,K_i})}[Y(i,l,S)=1].
\label{eq:psuf}
\end{equation}
The uniform measure gives a direct combinatorial interpretation: the number of successful subsets divided by the number of all $K_i$-element subsets. With the image, question, and intervention conditions fixed, $\psuf$ measures the abundance of visual combinations satisfying Eq.~\eqref{eq:success}. Hereafter, \emph{joint-preservation failure} means $Y=0$: at least one of the task-preservation, NLL, or KL conditions fails. This measures deviation from dense behavior; answer correctness is assessed separately by the task score against reference answers.

For each condition, two independent batches of $M=64$ samples yield
\begin{equation}
\widehat p^{(r)}_{i,l}=\frac1M\sum_{m=1}^{M}Y(i,l,S^{(r)}_m),
\qquad r\in\{1,2\}.
\end{equation}
Each intervention selects $K_i$ positions without replacement, and interventions are sampled independently. We pool the 128 observations for probability plots and retain the two batch estimates for error comparisons. Arranging these observations across layers and budgets produces the landscape.

\subsection{Probability differences and successful subsets}
\label{sec:successsets}
Differences in sufficiency probability constrain differences between sets of successful subsets. Fix an image $x$, layer $l$, and retention count $K$, and let two questions share the same indices of $N$ ordinary visual positions and the same joint-preservation criterion. Write $\Omega=\mathcal S_{N,K}$ for the subset space, $U$ for its uniform distribution, and $Y_q(S)=Y((x,q),l,S)$. The successful set for question $q$ is $A_q=\{S\in\Omega:Y_q(S)=1\}$, with sufficiency probability $p_q=U(A_q)$.

\begin{proposition}[Successful-subset disagreement across queries]
\label{prop:successdisagreement}
For any two questions $q_1,q_2$ under these conditions,
\begin{equation}
\begin{split}
|p_{q_1}-p_{q_2}|&\leq U(A_{q_1}\triangle A_{q_2})\\
&=\mathbb P_{S\sim U}[Y_{q_1}(S)\neq Y_{q_2}(S)].
\end{split}
\label{eq:successdisagreement}
\end{equation}
where $\triangle$ denotes the symmetric difference between the two successful sets.
\end{proposition}

The probability difference thus lower-bounds the fraction of retained subsets whose preservation outcomes differ between the two questions. Different sufficiency probabilities imply different members of the successful sets. This provides a set-level interpretation of within-image query differences, with each question evaluated using its own dense computational prefix and behavioral reference. The proof appears in Appendix~\ref{app:successdisagreement}. Appendix~\ref{app:selectionconcentration} further shows that, for $p_q>0$, $-\log p_q$ is the minimum KL divergence from the uniform distribution among selection distributions supported entirely on successful subsets. This characterizes the concentration of the successful region from the perspective of subset selection.

\subsection{Comparing landscapes against replication error}
\label{sec:distance}
Finite sampling produces variation even between curves measured under the same condition. We estimate this variation using independent replicates. For two curves on a common layer grid $\mathcal L$, we use the mean squared distance
\begin{equation}
 d(\mathbf a,\mathbf b)=\frac1{|\mathcal L|}
 \sum_{l\in\mathcal L}(a_l-b_l)^2,
\end{equation}
and compare between-input distances with within-input replication distances:
\begin{equation}
\begin{split}
\Delta_{\mathrm{curve}}={}&\operatorname{mean}_{i<j}
 d(\widehat{\mathbf p}^{(1)}_i,\widehat{\mathbf p}^{(1)}_j)\\
&-\operatorname{mean}_i
 d(\widehat{\mathbf p}^{(1)}_i,\widehat{\mathbf p}^{(2)}_i).
\end{split}
\label{eq:between}
\end{equation}
A positive value indicates that between-input variation exceeds the Monte Carlo replication error in this experiment.

To separate image identity from query conditions, we fix an image $x$ and compute over its three questions
\begin{equation}
\begin{split}
\Delta_Q(x)={}&\operatorname{mean}_{q<q'}
 d(\widehat{\mathbf p}^{(1)}_{x,q},\widehat{\mathbf p}^{(1)}_{x,q'})\\
&-\operatorname{mean}_q
 d(\widehat{\mathbf p}^{(1)}_{x,q},\widehat{\mathbf p}^{(2)}_{x,q}).
\end{split}
\label{eq:query}
\end{equation}
The overall $\Delta_Q$ gives equal weight to each image. This paired design directly measures within-image query differences. Variation in the length, answer format, and semantics of natural questions jointly constitutes the query condition.

To summarize where high sufficiency first appears with depth, we define the \emph{first high-sufficiency position} as the first observed layer with $\widehat p\geq0.80$. Cases that never reach the threshold by the final layer form a right-censored category. We refer to this discrete position as the boundary: it records the first threshold crossing without requiring subsequent observed layers to remain above the threshold. Subtracting the replication disagreement rate from the between-input boundary disagreement rate gives $\Delta_b$; the corresponding within-image statistic is $\Delta_{Q,b}$.
\section{Experimental Analysis}
\label{sec:experiments}
We test whether sufficiency landscapes reveal input variation beyond average depth trends. We first compare between-input differences with independent replication error, then hold images fixed to examine query conditions, and finally vary the retention budget to determine whether query differences persist across compression levels.

\subsection{Experimental setup}
\label{sec:setup}
Experiments cover Qwen2.5-VL-7B-Instruct~\cite{bai2025qwen}, InternVL2-8B~\cite{opengvlab2024internvl2}, and LLaVA-v1.5-7B~\cite{liu2024llava}, abbreviated as Qwen2.5-VL, InternVL2, and LLaVA. We evaluate GQA~\cite{hudson2019gqa} and TextVQA~\cite{singh2019textvqa}. For each model, the single-question experiment uses 32 images per task with one question per image. The within-image multi-question experiment uses another 32 GQA images with three questions each. Direct comparison of image IDs confirms that the two cohorts are disjoint. Inputs are paired across models.

The main retention ratio is $\rho=0.25$. The main curve grid spans L4--L24 at six layers for Qwen2.5-VL and L5--L30 at seven layers for InternVL2 and LLaVA. Cross-model effect comparisons use six relative-depth anchors. A budget supplement evaluates $\rho\in\{0.125,0.25,0.50\}$ on 12 preselected GQA images with three questions each. The controlled-selection prediction experiment in Section~\ref{sec:predictiveincrement} uses a seven-layer grid, specified separately from the main curve measurements here.

Each model completes 64 valid single-question inputs and 96 questions in the multi-question cohort; cohort construction is detailed in Appendix~\ref{app:provenance}. All intervals are percentile 95\% intervals from 5,000 image-cluster bootstrap resamples~\cite{field2007cluster}. Directional support is the proportion of resampled estimates that retain the stated direction.

\subsection{Average depth trends conceal input variation}
\label{sec:heterogeneity}
Table~\ref{tab:heterogeneity} shows clear variation between inputs under the same task and retention ratio. Net curve differences are positive for all three models on both GQA and TextVQA, with intervals entirely above zero and 100\% directional support. First high-sufficiency positions also differ beyond replication error. Inputs thus vary both in sufficiency probability and in the depth at which high sufficiency first appears.

\begin{table}[!htbp]
\centering\normalsize
\caption{Input heterogeneity at a retention ratio of 0.25. Each row uses 32 images; brackets indicate 95\% intervals. Net curve and boundary differences subtract replication error.}
\label{tab:heterogeneity}
\begin{tabular}{llcc}
\toprule
Model & Task & $\Delta_{\mathrm{curve}}$ & $\Delta_b$\\
\midrule
Qwen2.5-VL & GQA & $0.193\;[0.129,0.249]$ & $0.781\;[0.634,0.843]$\\
Qwen2.5-VL & TextVQA & $0.153\;[0.095,0.202]$ & $0.295\;[0.119,0.440]$\\
InternVL2 & GQA & $0.247\;[0.187,0.289]$ & $0.817\;[0.692,0.869]$\\
InternVL2 & TextVQA & $0.150\;[0.092,0.201]$ & $0.668\;[0.494,0.772]$\\
LLaVA & GQA & $0.221\;[0.161,0.268]$ & $0.773\;[0.627,0.844]$\\
LLaVA & TextVQA & $0.148\;[0.079,0.213]$ & $0.591\;[0.391,0.713]$\\
\bottomrule
\end{tabular}
\end{table}

\begin{figure}[!htbp]
\centering
\includegraphics[width=\linewidth]{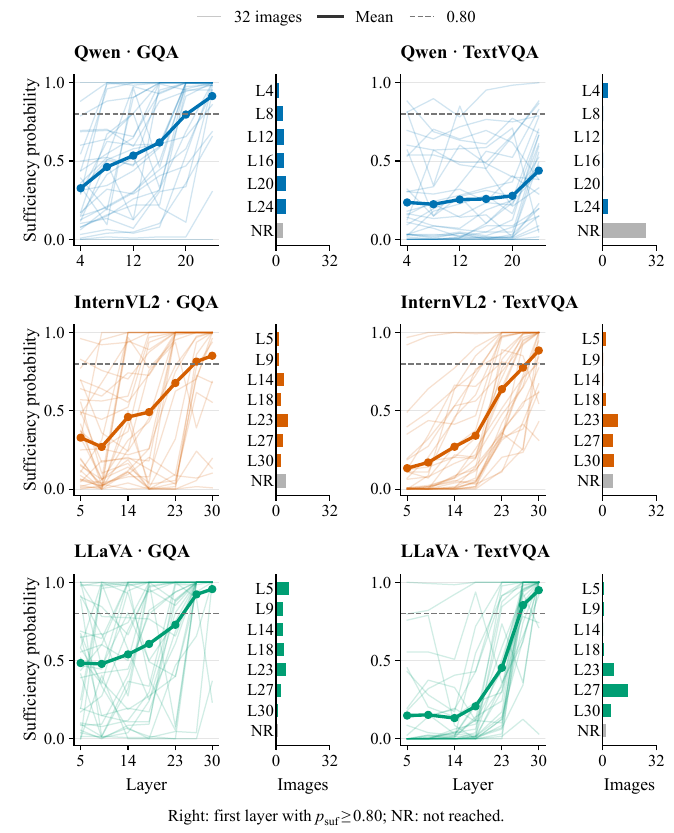}
\caption{Sufficiency landscapes for three models and two tasks at a retention ratio of 0.25. Rows correspond to models; columns show GQA and TextVQA. Light curves represent 32 images and dark curves their means, pooling two independent batches (128 interventions per condition). Lines connect only observed layers. The right side of each panel summarizes the first layer reaching 0.80 and retains cases that never reach the threshold.}
\label{fig:landscapes}
\end{figure}

Figure~\ref{fig:landscapes} juxtaposes the mean and all observed curves. Mean sufficiency probability increases with depth, but individual curves are widely dispersed, and most decrease between at least one pair of adjacent anchors. The average depth trend summarizes the overall direction, while input-specific curves characterize the conditions faced by individual inputs.

This dispersion also affects how well a task-level boundary represents individual inputs. For Qwen2.5-VL on TextVQA, the task-level boundary incurs an additional mean absolute error of 0.469 anchors, with an interval of $[-0.031,1.063]$; this estimate remains imprecise.

\subsection{Different queries change the landscape for the same image}
\label{sec:queryresults}
Between-input variation combines image and question changes. Holding the image fixed isolates differences associated with query conditions. In Table~\ref{tab:query}, all 32 images in each of the three models have point estimates satisfying $\Delta_Q(x)>0$, and the aggregate curve and boundary differences both have 100\% bootstrap directional support. Successful-subset abundance changes with the question even when visual content is fixed. This establishes query conditions as an important dimension of local prunability and connects them to the successful-set changes characterized by Proposition~\ref{prop:successdisagreement}.

\begin{table}[!htbp]
\centering\normalsize
\caption{Net within-image query differences. Each model uses 32 images with three questions per image; brackets indicate 95\% intervals.}
\label{tab:query}
\setlength{\tabcolsep}{3pt}
\begin{tabular}{lcc}
\toprule
Model & $\Delta_Q$ & $\Delta_{Q,b}$\\
\midrule
Qwen2.5-VL & $0.270\;[0.219,0.325]$ & $0.740\;[0.656,0.823]$\\
InternVL2 & $0.177\;[0.133,0.223]$ & $0.625\;[0.510,0.729]$\\
LLaVA & $0.177\;[0.146,0.209]$ & $0.740\;[0.646,0.823]$\\
\bottomrule
\end{tabular}
\end{table}

\begin{figure}[!htbp]
\centering
\includegraphics[width=\linewidth]{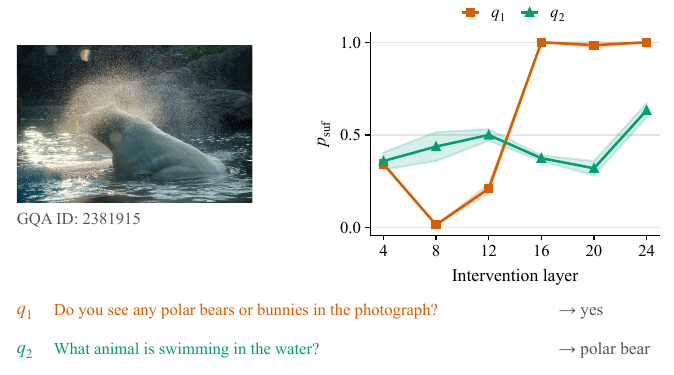}
\caption{Query-specific sufficiency curves for the same image. The photograph and corresponding Qwen2.5-VL curves appear side by side; the two questions and dataset answers (after arrows) appear below. The example is from GQA (ID: 2381915). Solid lines show pooled estimates, and shaded regions show the R1/R2 range, not confidence intervals.}
\label{fig:querypairs}
\end{figure}

For LLaVA, the image-level analysis yields positive net boundary differences for 31/32 images, and no single image determines the overall direction. Figure~\ref{fig:querypairs} illustrates two questions about the same image with distinct sufficiency probability curves.

\begin{figure}[!htbp]
\centering
\includegraphics[width=\linewidth]{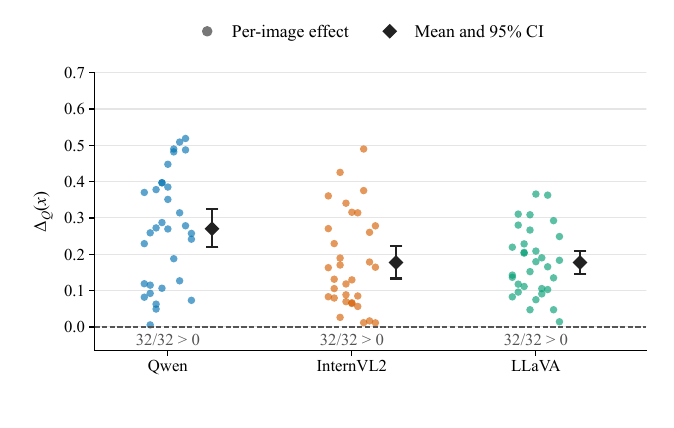}
\caption{Distribution of net query effects across images. Each model shows $\Delta_Q(x)$ for all 32 images. Colored dots are image-level net query curve differences; black diamonds and error bars show the equally weighted image mean and its 95\% interval. Dashed lines mark zero; all image-level point estimates are positive.}
\label{fig:queryeffects}
\end{figure}

Figure~\ref{fig:queryeffects} shows all image-level net query curve differences for the three models. Each model has $\Delta_Q(x)>0$ for all 32 images.

Comparing effect sizes across models requires a common depth reference. In the paired comparison at six relative-depth anchors, the difference in net curve effects between LLaVA and Qwen2.5-VL is $-0.0636$, with a 95\% interval of $[-0.1373,0.0106]$. Both models exhibit within-image query differences, but this interval does not clearly distinguish their magnitudes.

\FloatBarrier
\subsection{Query differences persist across budgets with model-dependent magnitudes}
\label{sec:budget}
Table~\ref{tab:budget} compares three retention ratios within the same 12-image supplementary cohort. All nine model--budget combinations have positive net query curve differences, with intervals above zero. Query-conditioned variation therefore persists across all three tested budgets.

\begin{table}[!htbp]
\centering\normalsize
\caption{Budget dependence of net query curve differences. Each model uses 12 images and 36 questions. Values are $\Delta_Q$ with 95\% intervals; Table~\ref{tab:query} reports the main 32-image results.}
\label{tab:budget}
\begin{tabular}{lccc}
\toprule
Model & Ratio 0.125 & Ratio 0.25 & Ratio 0.50\\
\midrule
Qwen2.5-VL & \makecell{$0.269$\\$[0.222,0.323]$} & \makecell{$0.315$\\$[0.268,0.364]$} & \makecell{$0.217$\\$[0.159,0.277]$}\\
InternVL2 & \makecell{$0.198$\\$[0.127,0.281]$} & \makecell{$0.185$\\$[0.106,0.272]$} & \makecell{$0.179$\\$[0.105,0.259]$}\\
LLaVA & \makecell{$0.248$\\$[0.186,0.312]$} & \makecell{$0.173$\\$[0.123,0.227]$} & \makecell{$0.165$\\$[0.115,0.217]$}\\
\bottomrule
\end{tabular}
\end{table}

The magnitude of query differences varies with budget in a model-dependent manner. For Qwen2.5-VL, the paired contrast between 0.50 and 0.25 is $-0.098$ (95\% CI: $[-0.172,-0.035]$). For LLaVA, the contrast between 0.125 and 0.25 is $+0.075$ (95\% CI: $[0.034,0.121]$). Both InternVL2 contrasts have intervals spanning zero. Within the tested range, increasing the retention budget does not produce a common pattern of effect-size change across models, while query differences remain present at each budget.

Criterion sensitivity provides further context. For LLaVA, requiring task preservation alone still yields a positive net query difference, but a smaller one than the joint criterion. Net boundary effects remain positive with thresholds of 0.7, 0.8, and 0.9 and with boundary definitions based on isotonic fitting or sustained preservation. The effect retains its direction when restricted to the 9 images for which the dense model answers all three questions correctly. Study-design details appear in Appendix~\ref{app:provenance}.
\section{Predictive Value of Sufficiency Landscapes}
\label{sec:prediction}
We connect successful-subset abundance to the risk of specific pruning strategies through incremental prediction for full methods, cross-image risk alerts, and generalization to new images under controlled selection rules. The prediction target is joint-preservation failure, $Z=1-Y(i,l,S)$, as defined in Eq.~\eqref{eq:success}. For each model, we independently fit paired predictors with and without sufficiency probability and measure the increment through the difference in log loss, giving equal weight to each image:
\begin{equation}
 \Delta\mathcal L=\mathcal L(F,\widetilde p_{\mathrm{suf}})-\mathcal L(F),
\end{equation}
Here, $F$ denotes the baseline features for the corresponding experiment, and negative values indicate improvement. AUC measures risk ranking; log loss and Brier score~\cite{brier1950verification,gneiting2007proper} measure probabilistic prediction quality. Intervals use 5,000 paired image-cluster bootstrap resamples. Comparisons are conditional on existing out-of-fold or fixed predictions, without refitting during resampling.

\subsection{Incremental failure prediction for full pruning methods}
\label{sec:fullmethods}
\label{sec:transplant}
On LLaVA/GQA, we run FastV~\cite{chen2024fastv}, VisionZip~\cite{yang2024visionzip}, and SparseVLM~\cite{zhang2025sparsevlm} using their native selection and generation pipelines. We also evaluate architecture-adapted configurations on Qwen2.5-VL and InternVL2. Baseline predictor B0 uses question type and summaries of dense outputs, full visual removal, and query similarity; B1 adds sufficiency probability. LLaVA uses the main cohort of 32 images with three questions each and eight-fold image-grouped logistic regression. All three methods share the sufficiency probability measured at layer 9 of base LLaVA. The architecture-adapted experiments retain the budgets and actions of their respective configurations, with 32 images per evaluation unit. Method configurations and predictor features appear in Appendix~\ref{app:fullmethods}.

Table~\ref{tab:fullmethods} compares log loss and AUC. Table~\ref{tab:transplantmetrics} adds Brier score, average precision (AP), and relative loss reductions to assess consistency across probability quality and risk ranking.

\begin{table}[!htbp]
\centering\normalsize
\caption{Incremental prediction for full pruning methods. B0 and B1 exclude and include sufficiency probability, respectively. Each row compares the two predictors within the same configuration.}
\label{tab:fullmethods}
\label{tab:transplantll}
\setlength{\tabcolsep}{5pt}
\begin{tabular}{llccc}
\toprule
Model & Method & \makecell{Log loss\\B0/B1 $\downarrow$} & \makecell{AUC\\B0/B1 $\uparrow$} & \makecell{$\Delta$ log loss\\95\% CI} \\
\midrule
LLaVA & FastV & \makecell{0.7578 /\\0.6703} & \makecell{0.5317 /\\0.6468} & $[-0.1613,-0.0144]$ \\
& VisionZip & \makecell{0.7005 /\\0.6649} & \makecell{0.5952 /\\0.6635} & $[-0.0767,+0.0064]$ \\
& SparseVLM & \makecell{0.6775 /\\0.6096} & \makecell{0.5938 /\\0.7339} & $[-0.1274,-0.0023]$ \\
\midrule
Qwen2.5-VL & FastV & \makecell{0.6240 /\\0.5408} & \makecell{0.7308 /\\0.8047} & $[-0.1523,-0.0124]$ \\
& VisionZip & \makecell{0.5269 /\\0.4837} & \makecell{0.7652 /\\0.8088} & $[-0.0806,-0.0059]$ \\
& SparseVLM & \makecell{0.6209 /\\0.4941} & \makecell{0.6974 /\\0.8185} & $[-0.2158,-0.0398]$ \\
\midrule
InternVL2 & FastV & \makecell{0.6503 /\\0.6496} & \makecell{0.7449 /\\0.7409} & $[-0.0025,+0.0012]$ \\
& VisionZip & \makecell{0.3224 /\\0.2317} & \makecell{0.6677 /\\0.8283} & $[-0.1536,-0.0364]$ \\
& SparseVLM & \makecell{0.6535 /\\0.5762} & \makecell{0.7298 /\\0.7854} & $[-0.1434,-0.0070]$ \\
\bottomrule
\end{tabular}
\end{table}

\begin{table}[!htbp]
\centering\normalsize
\caption{Multiple prediction metrics for full pruning methods. B0/B1 use the same paired configurations as Table~\ref{tab:fullmethods}. Lower Brier and higher AP are better. Relative reductions and nominal $p$ values refer to log loss.}
\label{tab:transplantmetrics}
\setlength{\tabcolsep}{5pt}
\begin{tabular}{llcccc}
\toprule
Model & Method & \makecell{Brier\\B0/B1 $\downarrow$} & \makecell{AP\\B0/B1 $\uparrow$} & \makecell{Relative\\reduction} & \makecell{Nominal\\$p$} \\
\midrule
LLaVA & FastV & \makecell{0.2695 /\\0.2313} & \makecell{0.5135 /\\0.5821} & 11.5\% & 0.026 \\
& VisionZip & \makecell{0.2514 /\\0.2344} & \makecell{0.6323 /\\0.6518} & 5.1\% & 0.105 \\
& SparseVLM & \makecell{0.2373 /\\0.2052} & \makecell{0.6093 /\\0.6652} & 10.0\% & 0.042 \\
\midrule
Qwen2.5-VL & FastV & \makecell{0.2138 /\\0.1830} & \makecell{0.8282 /\\0.8744} & 13.3\% & 0.031 \\
& VisionZip & \makecell{0.1744 /\\0.1587} & \makecell{0.8902 /\\0.9151} & 8.2\% & 0.030 \\
& SparseVLM & \makecell{0.2225 /\\0.1717} & \makecell{0.8131 /\\0.8910} & 20.4\% & 0.010 \\
\midrule
InternVL2 & FastV & \makecell{0.2307 /\\0.2305} & \makecell{0.6602 /\\0.6584} & 0.1\% & 0.476 \\
& VisionZip & \makecell{0.0892 /\\0.0653} & \makecell{0.9638 /\\0.9806} & 28.1\% & 0.006 \\
& SparseVLM & \makecell{0.2046 /\\0.1718} & \makecell{0.7681 /\\0.7640} & 11.8\% & 0.039 \\
\bottomrule
\end{tabular}
\end{table}

Together, the tables show predictive gains across methods and architectures. On LLaVA, AUC increases by 0.1151, 0.0683, and 0.1401 for the three methods, respectively, while log loss and Brier point estimates all decrease. Among the six architecture-adapted combinations, five reduce log loss by 8.2\%--28.1\%, with difference intervals below zero and improvements in both Brier and AUC. AP improves for four of these combinations; for InternVL2/SparseVLM, it decreases slightly from 0.7681 to 0.7640. The loss-difference interval for InternVL2/FastV spans zero, with small decreases in AUC and AP. Overall, successful-subset abundance provides risk information beyond existing input and model-behavior summaries across multiple pruning pipelines and architectures.

\subsection{Cross-image risk alerts}
\label{sec:methodtransfer}
We next test whether low sufficiency probability directly identifies high-risk questions. Each cohort contains 32 GQA images with three questions each; the transfer and main cohorts have disjoint image IDs. For each method, we flag the 32 questions with the lowest sufficiency probabilities. The risk difference (RD) is their joint-preservation failure rate minus that of the remaining 64 questions. Averaging equally across the three methods gives the macro risk difference. Image-cluster resampling uses common resampled records across methods and reranks questions to select the alert group.

\begin{table}[!htbp]
\centering\normalsize
\caption{Cross-image risk alerts. Each cohort has 96 questions, with 32 flagged per method. Macro RD averages equally across three methods; values and intervals are in percentage points. The transfer cohort tests risk concentration on a separate set of images.}
\label{tab:transplantrisk}
\setlength{\tabcolsep}{8pt}
\begin{tabular}{llcc}
\toprule
Model & Cohort & Macro RD & 95\% CI\\
\midrule
LLaVA & Transfer & 17.19 & [3.65, 30.21]\\
Qwen2.5-VL & Main & 41.67 & [29.69, 51.56]\\
 & Transfer & 38.54 & [27.60, 48.44]\\
InternVL2 & Main & 23.44 & [13.54, 34.38]\\
 & Transfer & 23.44 & [9.90, 33.85]\\
\bottomrule
\end{tabular}
\end{table}

In Table~\ref{tab:transplantrisk}, macro risk-difference intervals are above zero for all three models on the transfer cohort. Low sufficiency probability thus identifies questions with more concentrated failures on new images. Qwen2.5-VL and InternVL2 show the same direction in the main cohort. In the method-level LLaVA analysis, risk concentration is strongest for SparseVLM, while the FastV and VisionZip intervals span zero (Appendix Table~\ref{tab:fullalert}). Sufficiency measurements therefore provide a direct basis for alerts with a fixed quota.

\FloatBarrier
\subsection{Controlled selection rules and generalization to new images}
\label{sec:predictiveincrement}
\label{sec:newcohort}
To distinguish selection-rule effects from native method configurations, we fix the input, layer, and budget and compare three rules based on query similarity, feature distance, and central position. Each model uses 32 images, three questions per image, seven layers, and three selectors, yielding 2,016 records. Baseline features cover structural conditions, dense outputs, available single-token information summaries, query similarity, and the seven-layer full-visual-removal trajectory. The augmented predictor adds smoothed sufficiency probability estimated through independent random interventions. Layer-aware modeling uses image-grouped nested cross-validation with eight outer folds and four inner folds~\cite{varma2006cv}. Baseline loss selects the learner family, and the two predictors are tuned separately. Selection rules, features, and fitting details appear in Appendix~\ref{app:predictors}.

Table~\ref{tab:prediction} compares out-of-fold evaluation on the original cohort with evaluation of fixed predictors on new images. On the original cohort, log loss point estimates decrease for all three models. Difference intervals are below zero for InternVL2 and LLaVA and span zero for Qwen2.5-VL. We then freeze predictors trained on all 32 original images and apply them directly to another 32 GQA images, each with one attribute, one relation, and one object question. On these new images, log loss decreases by 9.0\%, 13.0\%, and 14.1\%, respectively, with all difference intervals below zero. Brier differences are also negative with intervals excluding zero (Appendix Table~\ref{tab:brier}).

\begin{table}[!htbp]
\centering\normalsize
\caption{Incremental prediction under controlled selection rules and generalization to new images. All results use layer-aware modeling. Differences subtract baseline log loss from log loss with sufficiency probability; negative values are favorable. New-image evaluation freezes predictors trained on the original cohort.}
\label{tab:prediction}
\setlength{\tabcolsep}{6pt}
\begin{tabular}{lcccc}
\toprule
Model & \makecell{Without\\$\psuf$} & \makecell{With\\$\psuf$} & \makecell{$\Delta\mathcal L$\\{}[95\% CI]} & \makecell{Relative\\reduction}\\
\midrule
\multicolumn{5}{l}{Original cohort: image-grouped out-of-fold evaluation}\\
Qwen2.5-VL & 0.4288 & 0.4116 & \makecell{$-0.0172$\\$[-0.0348,0.0017]$} & 4.0\%\\
InternVL2 & 0.4185 & 0.3513 & \makecell{$-0.0672$\\$[-0.1012,-0.0398]$} & 16.1\%\\
LLaVA & 0.4811 & 0.4014 & \makecell{$-0.0797$\\$[-0.1069,-0.0537]$} & 16.6\%\\
\midrule
\multicolumn{5}{l}{New images: direct evaluation of fixed predictors}\\
Qwen2.5-VL & 0.4092 & 0.3724 & \makecell{$-0.0367$\\$[-0.0480,-0.0259]$} & 9.0\%\\
InternVL2 & 0.3935 & 0.3424 & \makecell{$-0.0511$\\$[-0.0692,-0.0325]$} & 13.0\%\\
LLaVA & 0.4635 & 0.3982 & \makecell{$-0.0654$\\$[-0.0838,-0.0473]$} & 14.1\%\\
\bottomrule
\end{tabular}
\end{table}

\begin{figure}[!htbp]
\centering
\includegraphics[width=\linewidth]{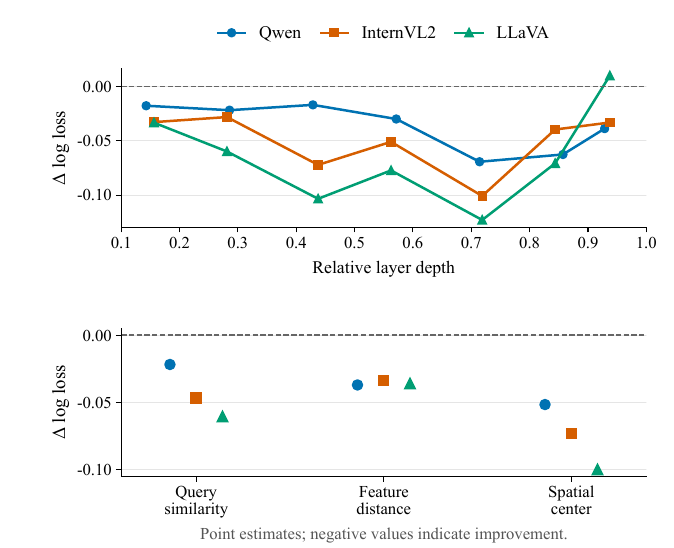}
\caption{Incremental predictive value of sufficiency probability on new images. The upper panel shows results by relative layer position; the lower panel compares the three controlled selection rules. Points show log loss with sufficiency probability minus baseline log loss; negative values indicate improvement.}
\label{fig:prediction}
\end{figure}

On new images, log loss improves for every selector, at all seven layers for Qwen2.5-VL and InternVL2, and at six layers for LLaVA (Fig.~\ref{fig:prediction}). Appendix~\ref{app:predictionresults} reports fixed logistic regression, feature ablations, and new-cohort refitting. Together with full-method results, these findings connect successful-subset abundance to pruning risk.
\section{Discussion and Limitations}
\label{sec:discussion}
\paragraph{Interpretation of the measurement}
Visual sufficiency extends the analysis of pruning reliability from individual retained sets to the abundance of successful combinations. It measures preservation of dense behavior after a given computational prefix and can therefore preserve an incorrect dense answer. The joint event also captures probabilistic changes beyond the generated answer. In the main-cohort experiments with full methods, 19, 26, and 24 questions satisfy task preservation under FastV, VisionZip, and SparseVLM, respectively, but fail the NLL or KL condition (Appendix~\ref{app:methoddiagnostics}). Answer preservation and joint preservation thus reveal different aspects of behavioral change after pruning.

\paragraph{Query conditions and landscape variation}
The within-image results show that visual content alone does not determine the sufficiency landscape. Query conditions encompass jointly varying factors such as length, semantics, referents, and answer format. Local decreases in individual curves are observed under finite sampling and a discrete layer grid. The first high-sufficiency position summarizes threshold attainment, while the full curve retains more information about variation with depth.

\paragraph{Scope of the predictive results}
The full-method experiments show that one sufficiency probability measured on base LLaVA provides additional risk information for three native pruning pipelines. Experiments across architectures reproduce this increment in several adapted configurations on Qwen2.5-VL and InternVL2. Under controlled selection rules, predictors are fitted independently within each model, and evaluation of fixed predictors on new GQA images supports generalization across images. Together, these experiments establish an empirical link between successful-subset abundance and specific pruning risks. The controlled central-position rule uses a proxy grid. Predictive gains on relation questions also suggest that joint preservation of multiple visual elements may contribute information; testing this mechanism requires direct analysis of joint responses.

\paragraph{From offline measurement to pruning decisions}
Sufficiency landscapes provide a measurable target for risk-aware pruning: identifying inputs with few successful combinations that require more careful subset selection. The current measurement runs 128 random interventions per condition and uses reference answers to compute NLL and KL, making it suitable for offline analysis. Turning it into an online decision signal requires inexpensive sufficiency estimators and evaluation signals available at inference time. It also requires jointly evaluating task performance and total computational cost when alerts trigger fallback inference or budget adjustments.
\section{Conclusion}
\label{sec:conclusion}
Visual sufficiency landscapes characterize the flexibility in selecting effective retained sets through the abundance of successful combinations. This provides a combinatorial perspective on the conditions under which different pruning strategies can succeed. Experiments on three vision-language models show that landscape differences across inputs and across questions about the same image exceed independent replication error, with query differences persisting across multiple budgets. Sufficiency probability also adds information beyond existing features for predicting joint-preservation failure under full pruning methods and controlled selection rules. Replication across architectures, direct evaluation on new images, and cross-image risk alerts support the practical value of this measurement. Successful-combination abundance and its query dependence are therefore important dimensions of visual token pruning reliability and provide a basis for risk-aware pruning design.
\section*{Declaration of AI-assisted code development}
During the development of the experimental code, the authors used AI tools to assist with coding. All code was reviewed by the authors, who take full responsibility for the code and its use in this study.

\FloatBarrier
\bibliographystyle{elsarticle-num-compact}
\bibliography{references}

@article{he2026coreprune,
 author={He, Shengli and Liang, Yongchao and He, Roumeng and Zeng, Junjie and He, Jiyuan and Fang, Xin and Wu, Can and Zheng, Li},
 title={From Token Importance to Conditional Removability: Rethinking Visual Token Pruning in Multimodal Large Language Models},
 journal={arXiv preprint arXiv:2609.26484}, year={2026}, doi={10.48550/arXiv.2609.26484}, url={https://arxiv.org/abs/2609.26484}}

@inproceedings{endo2025feather,
 author={Endo, Mark and Wang, Xiaohan and Yeung-Levy, Serena},
 title={Feather the Throttle: Revisiting Visual Token Pruning for Vision-Language Model Acceleration},
 booktitle={Proceedings of the IEEE/CVF International Conference on Computer Vision}, year={2025}, pages={22826--22835}, url={https://openaccess.thecvf.com/content/ICCV2025/html/Endo_Feather_the_Throttle_Revisiting_Visual_Token_Pruning_for_Vision-Language_Model_ICCV_2025_paper.html}}

@inproceedings{chen2024fastv,
 author={Chen, Liang and Zhao, Haozhe and Liu, Tianyu and Bai, Shuai and Lin, Junyang and Zhou, Chang and Chang, Baobao},
 title={An Image is Worth 1/2 Tokens After Layer 2: Plug-and-Play Inference Acceleration for Large Vision-Language Models},
 booktitle={European Conference on Computer Vision (ECCV)}, year={2024}, pages={19--35}, doi={10.1007/978-3-031-73004-7_2}, url={https://doi.org/10.1007/978-3-031-73004-7_2}}

@inproceedings{yang2024visionzip,
 author={Yang, Senqiao and Chen, Yukang and Tian, Zhuotao and Wang, Chengyao and Li, Jingyao and Yu, Bei and Jia, Jiaya},
 title={{VisionZip}: Longer is Better but Not Necessary in Vision Language Models},
 booktitle={Proceedings of the IEEE/CVF Conference on Computer Vision and Pattern Recognition}, year={2025}, pages={19792--19802}, url={https://openaccess.thecvf.com/content/CVPR2025/html/Yang_VisionZip_Longer_is_Better_but_Not_Necessary_in_Vision_Language_CVPR_2025_paper.html}}

@inproceedings{alvar2025divprune,
 author={Alvar, Saeed Ranjbar and Singh, Gursimran and Akbari, Mohammad and Zhang, Yong},
 title={{DivPrune}: Diversity-based Visual Token Pruning for Large Multimodal Models},
 booktitle={Proceedings of the IEEE/CVF Conference on Computer Vision and Pattern Recognition}, year={2025}, pages={9392--9401}, url={https://openaccess.thecvf.com/content/CVPR2025/html/Alvar_DivPrune_Diversity-based_Visual_Token_Pruning_for_Large_Multimodal_Models_CVPR_2025_paper.html}}

@inproceedings{zhang2025sparsevlm,
 author={Zhang, Yuan and Fan, Chun-Kai and Ma, Junpeng and Zheng, Wenzhao and Huang, Tao and Cheng, Kuan and Gudovskiy, Denis A. and Okuno, Tomoyuki and Nakata, Yohei and Keutzer, Kurt and Zhang, Shanghang},
 title={{SparseVLM}: Visual Token Sparsification for Efficient Vision-Language Model Inference},
 booktitle={Proceedings of the 42nd International Conference on Machine Learning}, series={Proceedings of Machine Learning Research}, volume={267}, pages={74840--74857}, publisher={PMLR}, year={2025}, url={https://proceedings.mlr.press/v267/zhang25s.html}}

@inproceedings{ye2025atp,
 author={Ye, Xubing and Gan, Yukang and Ge, Yixiao and Zhang, Xiao-Ping and Tang, Yansong},
 title={{ATP-LLaVA}: Adaptive Token Pruning for Large Vision Language Models},
 booktitle={Proceedings of the IEEE/CVF Conference on Computer Vision and Pattern Recognition}, year={2025}, pages={24972--24982}, url={https://openaccess.thecvf.com/content/CVPR2025/html/Ye_ATP-LLaVA_Adaptive_Token_Pruning_for_Large_Vision_Language_Models_CVPR_2025_paper.html}}

@inproceedings{guo2025crop,
 author={Guo, Jiawei and Zhai, Feifei and Jian, Pu and Wei, Qianrun and Zhou, Yu},
 title={{CROP}: Contextual Region-Oriented Visual Token Pruning},
 booktitle={Proceedings of the 2025 Conference on Empirical Methods in Natural Language Processing},
 year={2025}, pages={9756--9772}, doi={10.18653/v1/2025.emnlp-main.492}, url={https://aclanthology.org/2025.emnlp-main.492/}}

@inproceedings{wang2026horizon,
 author={Wang, Yahong and Wu, Juncheng and Ni, Zhangkai and Yang, Longzhen and Liu, Yihang and Yang, Chengmei and Wen, Ying and He, Lianghua and Tang, Xianfeng and Liu, Hui and Zhou, Yuyin},
 title={When Token Pruning is Worse than Random: Understanding Visual Token Information in {VLLMs}},
 booktitle={Proceedings of the IEEE/CVF Conference on Computer Vision and Pattern Recognition},
 year={2026}, pages={31910--31919}, url={https://openaccess.thecvf.com/content/CVPR2026/html/Wang_When_Token_Pruning_is_Worse_than_Random_Understanding_Visual_Token_CVPR_2026_paper.html}}

@article{li2026occamtoken,
 author={Li, Geng and Chen, Guohao and Chen, Ting and Shan, Shilin and Zuo, Kuangji and Lyu, Bofan and An, Tuo and Li, Gen and Yang, Jianfei},
 title={{OccamToken}: Efficient {VLM} Inference with Training-Free and Budget-Adaptive Token Pruning},
 journal={arXiv preprint arXiv:2605.29657}, year={2026}, doi={10.48550/arXiv.2605.29657}, url={https://arxiv.org/abs/2605.29657}}

@article{he2026qcpruner,
 author={He, Shengli and Liang, Yongchao and He, Roumeng and Zeng, Junjie and He, Jiyuan and Wu, Can and Zheng, Li},
 title={{QCPruner}: Query-Conditioned Population Coverage for Visual Token Pruning},
 journal={arXiv preprint arXiv:2609.19990}, year={2026}, doi={10.48550/arXiv.2609.19990}, url={https://arxiv.org/abs/2609.19990}}

@inproceedings{hudson2019gqa,
 author={Hudson, Drew A. and Manning, Christopher D.},
 title={{GQA}: A New Dataset for Real-World Visual Reasoning and Compositional Question Answering},
 booktitle={Proceedings of the IEEE/CVF Conference on Computer Vision and Pattern Recognition}, year={2019}, pages={6693--6702}, doi={10.1109/CVPR.2019.00686},
 url={https://openaccess.thecvf.com/content_CVPR_2019/html/Hudson_GQA_A_New_Dataset_for_Real-World_Visual_Reasoning_and_Compositional_CVPR_2019_paper.html}}

@inproceedings{singh2019textvqa,
 author={Singh, Amanpreet and Natarajan, Vivek and Shah, Meet and Jiang, Yu and Chen, Xinlei and Batra, Dhruv and Parikh, Devi and Rohrbach, Marcus},
 title={Towards {VQA} Models That Can Read},
 booktitle={Proceedings of the IEEE/CVF Conference on Computer Vision and Pattern Recognition}, year={2019}, pages={8317--8326}, url={https://openaccess.thecvf.com/content_CVPR_2019/html/Singh_Towards_VQA_Models_That_Can_Read_CVPR_2019_paper.html}}

@article{bai2025qwen,
 author={Bai, Shuai and Chen, Keqin and Liu, Xuejing and Wang, Jialin and Ge, Wenbin and Song, Sibo and Dang, Kai and Wang, Peng and Wang, Shijie and Tang, Jun and Zhong, Humen and Zhu, Yuanzhi and Yang, Mingkun and Li, Zhaohai and Wan, Jianqiang and Wang, Pengfei and Ding, Wei and Fu, Zheren and Xu, Yiheng and Ye, Jiabo and Zhang, Xi and Xie, Tianbao and Cheng, Zesen and Zhang, Hang and Yang, Zhibo and Xu, Haiyang and Lin, Junyang},
 title={{Qwen2.5-VL} Technical Report}, journal={arXiv preprint arXiv:2502.13923}, year={2025}, doi={10.48550/arXiv.2502.13923}, url={https://arxiv.org/abs/2502.13923}}

@misc{opengvlab2024internvl2,
 author={{OpenGVLab Team}},
 title={{InternVL2}: Better than the Best---Expanding Performance Boundaries of Open-Source Multimodal Models with the Progressive Scaling Strategy},
 year={2024}, howpublished={Project report}, note={Accessed 15 October 2026}, url={https://internvl.github.io/blog/2024-07-02-InternVL-2.0/}}

@inproceedings{liu2024llava,
 author={Liu, Haotian and Li, Chunyuan and Li, Yuheng and Lee, Yong Jae},
 title={Improved Baselines with Visual Instruction Tuning},
 booktitle={Proceedings of the IEEE/CVF Conference on Computer Vision and Pattern Recognition}, year={2024}, pages={26296--26306}, url={https://openaccess.thecvf.com/content/CVPR2024/html/Liu_Improved_Baselines_with_Visual_Instruction_Tuning_CVPR_2024_paper.html}}

@inproceedings{mao2016refcocog,
 author={Mao, Junhua and Huang, Jonathan and Toshev, Alexander and Camburu, Oana and Yuille, Alan L. and Murphy, Kevin},
 title={Generation and Comprehension of Unambiguous Object Descriptions},
 booktitle={Proceedings of the IEEE Conference on Computer Vision and Pattern Recognition},
 year={2016}, pages={11--20}, url={https://openaccess.thecvf.com/content_cvpr_2016/html/Mao_Generation_and_Comprehension_CVPR_2016_paper.html}}

@article{brier1950verification,
 author={Brier, Glenn W.},
 title={Verification of Forecasts Expressed in Terms of Probability},
 journal={Mon. Weather Rev.}, volume={78}, number={1}, pages={1--3}, year={1950},
 doi={10.1175/1520-0493(1950)078<0001:VOFEIT>2.0.CO;2},
 url={https://journals.ametsoc.org/view/journals/mwre/78/1/1520-0493_1950_078_0001_vofeit_2_0_co_2.xml}}

@article{gneiting2007proper,
 author={Gneiting, Tilmann and Raftery, Adrian E.},
 title={Strictly Proper Scoring Rules, Prediction, and Estimation},
 journal={J. Am. Stat. Assoc.}, volume={102}, number={477}, pages={359--378}, year={2007},
 doi={10.1198/016214506000001437}, url={https://doi.org/10.1198/016214506000001437}}

@article{field2007cluster,
 author={Field, C. A. and Welsh, A. H.},
 title={Bootstrapping Clustered Data},
 journal={J. R. Stat. Soc. Ser. B Stat. Methodol.},
 volume={69}, number={3}, pages={369--390}, year={2007},
 doi={10.1111/j.1467-9868.2007.00593.x}, url={https://rss.onlinelibrary.wiley.com/doi/abs/10.1111/j.1467-9868.2007.00593.x}}

@article{varma2006cv,
 author={Varma, Sudhir and Simon, Richard},
 title={Bias in Error Estimation When Using Cross-Validation for Model Selection},
 journal={BMC Bioinformatics}, volume={7}, pages={91}, year={2006},
 doi={10.1186/1471-2105-7-91}, url={https://link.springer.com/article/10.1186/1471-2105-7-91}}

@article{pedregosa2011sklearn,
 author={Pedregosa, Fabian and Varoquaux, Ga{\"e}l and Gramfort, Alexandre and Michel, Vincent and Thirion, Bertrand and Grisel, Olivier and Blondel, Mathieu and Prettenhofer, Peter and Weiss, Ron and Dubourg, Vincent and Vanderplas, Jake and Passos, Alexandre and Cournapeau, David and Brucher, Matthieu and Perrot, Matthieu and Duchesnay, {\'E}douard},
 title={Scikit-learn: Machine Learning in {Python}},
 journal={J. Mach. Learn. Res.}, volume={12}, number={85}, pages={2825--2830}, year={2011}, url={https://jmlr.org/papers/v12/pedregosa11a.html}}

@inproceedings{bolya2023tome,
 author={Bolya, Daniel and Fu, Cheng-Yang and Dai, Xiaoliang and Zhang, Peizhao and Feichtenhofer, Christoph and Hoffman, Judy},
 title={Token Merging: Your {ViT} But Faster},
 booktitle={International Conference on Learning Representations},
 year={2023}, url={https://openreview.net/forum?id=JroZRaRw7Eu}}

@inproceedings{cao2023pumer,
 author={Cao, Qingqing and Paranjape, Bhargavi and Hajishirzi, Hannaneh},
 title={{PuMer}: Pruning and Merging Tokens for Efficient Vision Language Models},
 booktitle={Proceedings of the 61st Annual Meeting of the Association for Computational Linguistics (Volume 1: Long Papers)},
 year={2023}, pages={12890--12903}, doi={10.18653/v1/2023.acl-long.721},
 url={https://aclanthology.org/2023.acl-long.721/}}

@inproceedings{shang2025prumerge,
 author={Shang, Yuzhang and Cai, Mu and Xu, Bingxin and Lee, Yong Jae and Yan, Yan},
 title={{LLaVA-PruMerge}: Adaptive Token Reduction for Efficient Large Multimodal Models},
 booktitle={Proceedings of the IEEE/CVF International Conference on Computer Vision},
 year={2025}, pages={22857--22867},
 url={https://openaccess.thecvf.com/content/ICCV2025/html/Shang_LLaVA-PruMerge_Adaptive_Token_Reduction_for_Efficient_Large_Multimodal_Models_ICCV_2025_paper.html}}

@inproceedings{song2025trim,
 author={Song, Dingjie and Wang, Wenjun and Chen, Shunian and Wang, Xidong and Guan, Michael X. and Wang, Benyou},
 title={Less is More: A Simple yet Effective Token Reduction Method for Efficient Multi-modal {LLMs}},
 booktitle={Proceedings of the 31st International Conference on Computational Linguistics},
 year={2025}, pages={7614--7623}, url={https://aclanthology.org/2025.coling-main.508/}}

@inproceedings{ye2025fitprune,
 author={Ye, Weihao and Wu, Qiong and Lin, Wenhao and Zhou, Yiyi},
 title={Fit and Prune: Fast and Training-free Visual Token Pruning for Multi-modal Large Language Models},
 booktitle={Proceedings of the AAAI Conference on Artificial Intelligence},
 year={2025}, volume={39}, pages={22128--22136},
 doi={10.1609/aaai.v39i21.34366}, url={https://ojs.aaai.org/index.php/AAAI/article/view/34366}}

@inproceedings{yang2025topv,
 author={Yang, Cheng and Sui, Yang and Xiao, Jinqi and Huang, Lingyi and Gong, Yu and Li, Chendi and Yan, Jinghua and Bai, Yu and Sadayappan, Ponnuswamy and Hu, Xia and Yuan, Bo},
 title={{TopV}: Compatible Token Pruning with Inference Time Optimization for Fast and Low-Memory Multimodal Vision Language Model},
 booktitle={Proceedings of the IEEE/CVF Conference on Computer Vision and Pattern Recognition},
 year={2025}, pages={19803--19813},
 url={https://openaccess.thecvf.com/content/CVPR2025/html/Yang_TopV_Compatible_Token_Pruning_with_Inference_Time_Optimization_for_Fast_CVPR_2025_paper.html}}

@inproceedings{zhang2025cdpruner,
 author={Zhang, Qizhe and Liu, Mengzhen and Li, Lichen and Lu, Ming and Zhang, Yuan and Pan, Junwen and She, Qi and Zhang, Shanghang},
 title={Beyond Attention or Similarity: Maximizing Conditional Diversity for Token Pruning in {MLLMs}},
 booktitle={Advances in Neural Information Processing Systems},
 year={2025}, volume={38}, pages={28761--28791}, doi={10.52202/085713-0855},
 url={https://doi.org/10.52202/085713-0855}}

@inproceedings{zhu2026hawk,
 author={Zhu, Qihui and Zhang, Tao and Wang, Yuchen and Chen, Shuangwu and Tan, Xiaobin and Yang, Jian and Liu, Yang and Pan, Yinfei},
 title={{HAWK}: Head Importance-Aware Visual Token Pruning in Multimodal Models},
 booktitle={Proceedings of the IEEE/CVF Conference on Computer Vision and Pattern Recognition},
 year={2026}, pages={39583--39592}, url={https://openaccess.thecvf.com/content/CVPR2026/html/Zhu_HAWK_Head_Importance-Aware_Visual_Token_Pruning_in_Multimodal_Models_CVPR_2026_paper.html}}

@inproceedings{lin2025vtw,
 author={Lin, Zhihang and Lin, Mingbao and Lin, Luxi and Ji, Rongrong},
 title={Boosting Multimodal Large Language Models with Visual Tokens Withdrawal for Rapid Inference},
 booktitle={Proceedings of the AAAI Conference on Artificial Intelligence},
 year={2025}, volume={39}, pages={5334--5342},
 doi={10.1609/aaai.v39i5.32567}, url={https://ojs.aaai.org/index.php/AAAI/article/view/32567}}

@inproceedings{xing2025pyramiddrop,
 author={Xing, Long and Huang, Qidong and Dong, Xiaoyi and Lu, Jiajie and Zhang, Pan and Zang, Yuhang and Cao, Yuhang and He, Conghui and Wang, Jiaqi and Wu, Feng and Lin, Dahua},
 title={Conical Visual Concentration for Efficient Large Vision-Language Models},
 booktitle={Proceedings of the IEEE/CVF Conference on Computer Vision and Pattern Recognition},
 year={2025}, pages={14593--14603},
 url={https://openaccess.thecvf.com/content/CVPR2025/html/Xing_Conical_Visual_Concentration_for_Efficient_Large_Vision-Language_Models_CVPR_2025_paper.html}}

@inproceedings{fan2026embedlens,
 author={Fan, Yingqi and Tong, Junlong and Zhao, Anhao and Shen, Xiaoyu},
 title={What Do Visual Tokens Really Encode? Uncovering Sparsity and Redundancy in Multimodal Large Language Models},
 booktitle={Proceedings of the IEEE/CVF Conference on Computer Vision and Pattern Recognition},
 year={2026}, pages={11987--11997}, url={https://openaccess.thecvf.com/content/CVPR2026/html/Fan_What_Do_Visual_Tokens_Really_Encode_Uncovering_Sparsity_and_Redundancy_CVPR_2026_paper.html}}

@inproceedings{kim2026dstp,
 author={Kim, Jiwan and Kim, Kibum and Kim, Wonjoong and Lee, Byung-Kwan and Park, Chanyoung},
 title={Why and When Visual Token Pruning Fails? A Study on Relevant Visual Information Shift in {MLLMs} Decoding},
 booktitle={Computer Vision -- ECCV 2026},
 publisher={Springer Nature Switzerland}, address={Cham}, year={2026}, pages={244--263},
 doi={10.1007/978-3-032-37281-9_15}, url={https://doi.org/10.1007/978-3-032-37281-9_15}}

@inproceedings{deyoung2020eraser,
 author={DeYoung, Jay and Jain, Sarthak and Rajani, Nazneen Fatema and Lehman, Eric and Xiong, Caiming and Socher, Richard and Wallace, Byron C.},
 title={{ERASER}: A Benchmark to Evaluate Rationalized {NLP} Models},
 booktitle={Proceedings of the 58th Annual Meeting of the Association for Computational Linguistics},
 year={2020}, pages={4443--4458}, doi={10.18653/v1/2020.acl-main.408},
 url={https://aclanthology.org/2020.acl-main.408/}}

@inproceedings{kamp2025rationales,
 author={Kamp, Jonathan and Beinborn, Lisa and Fokkens, Antske},
 title={Learning from Sufficient Rationales: Analysing the Relationship Between Explanation Faithfulness and Token-level Regularisation Strategies},
 booktitle={Proceedings of the 14th International Joint Conference on Natural Language Processing and the 4th Conference of the Asia-Pacific Chapter of the Association for Computational Linguistics},
 year={2025}, pages={3028--3044}, doi={10.18653/v1/2025.ijcnlp-long.162},
 url={https://aclanthology.org/2025.ijcnlp-long.162/}}

@inproceedings{guo2017calibration,
 author={Guo, Chuan and Pleiss, Geoff and Sun, Yu and Weinberger, Kilian Q.},
 title={On Calibration of Modern Neural Networks},
 booktitle={Proceedings of the 34th International Conference on Machine Learning},
 series={Proceedings of Machine Learning Research}, volume={70}, pages={1321--1330},
 year={2017}, publisher={PMLR}, url={https://proceedings.mlr.press/v70/guo17a.html}}

@article{tan2026confidence,
 author={Tan, Kaizhen and Feng, Yang and Du, Heqing and Hong, Hanzhe and Tao, Siru and Xu, Xin},
 title={Confidence under Visual Token Pruning: Removed Evidence and Risk-Controlled Token Budgets for {MLLMs}},
 journal={arXiv preprint arXiv:2604.12035v4}, year={2026}, doi={10.48550/arXiv.2604.12035},
 url={https://arxiv.org/abs/2604.12035v4}}
\label{page:mainend}

\clearpage
\appendix
\renewcommand{\thesection}{\Alph{section}}
\counterwithin{table}{section}
\counterwithin{figure}{section}
\renewcommand{\thetable}{\Alph{section}.\arabic{table}}
\renewcommand{\thefigure}{\Alph{section}.\arabic{figure}}
\section{Implementation, Predictor Features, and Evaluation Details}
\label{app:predictors}
\subsection{Model and task implementation}
GQA~\cite{hudson2019gqa} uses the validation balanced instructions and corresponding images; TextVQA~\cite{singh2019textvqa} uses the validation split. Both tasks request a single word or short phrase, with a maximum of 16 generated tokens, stochastic generation disabled, and the KV cache disabled. The common prompt is ``Answer the question using a single word or short phrase. Question: \{question\} Answer:, with line breaks and model chat templates preserved by each adapter. Qwen2.5-VL and InternVL2 use bfloat16, and LLaVA uses float16. Language decoders use eager attention. The input pixel range for Qwen2.5-VL is 200704--1003520. InternVL2 uses at most 12 dynamic image tiles plus a thumbnail; LLaVA uses padding-based preprocessing and 576 visual tokens. Cross-model comparisons retain each model	extquotesingle s native preprocessing.

The two original sampling seeds are 2026091711 and 2026091712; the bootstrap seed for the main landscape analyses is 2026091714. Each intervention removes only ordinary visual positions at the current layer, with subsequent computation following the model-specific physical compression pipeline. Special boundary tokens are excluded from the sampling pool. Engineering checks before measurement cover dense inference, full-retention and repeated-execution consistency, visual-position ranges, physical subset interventions, and task parsing. Seeds, sampling lists, and statistical configurations for supplementary stages are recorded with their run outputs.

\subsection{Controlled selection rules and probability features}
The three controlled rules select by query similarity, feature distance, and central position. Query similarity uses the cosine similarity between the final prompt-position representation and visual representations. Feature distance uses Euclidean distance from the mean visual representation. Central position uses a row-major proxy grid constructed from the token count and aspect ratio. For the subset $S_a$ produced by selector $a$, sufficiency probability is estimated using independent random interventions and smoothed as
\begin{equation}
 \widetilde p_{\mathrm{suf}}=\frac{n_{\mathrm{success}}+0.5}{129}.
\end{equation}
Fixed logistic regression and layer-aware modeling use identical image groups and selector labels. They assess incremental information under a fixed linear mapping and a richer representation of layer effects, respectively.

\subsection{Features and coverage}
The feature ablation uses four nested L2-regularized logistic regression models. $M_0$ includes relative layer position, retention ratio, visual token count, actual retention count, question length in characters, dense reference-answer NLL, dense output entropy, and the gap between the top two output probabilities. $M_1$ adds task preservation, NLL increase, KL, and joint preservation under full visual removal at the same layer, together with the mean, standard deviation, mean absolute value, near-zero fraction, and positive and negative fractions of single-token information increments. $M_2$ adds the mean and standard deviation of cosine similarities between the common query representation and visual representations, as well as normalized entropy, top-$K$ probability mass, and the Gini coefficient after softmax. $M_3$ adds only smoothed sufficiency probability. All four include selector type.

The common query representation is the hidden state at the final prompt position at the intervention boundary. Following Information Horizon~\cite{wang2026horizon}, single-token information compares single-position retention with full visual removal through changes in target-output probability. Summaries are computed for 252 conditions spanning 12 preselected images, three questions each, and seven layers, using at most 256 deterministically equispaced positions per condition. All 252 Qwen2.5-VL conditions in the original cohort use sampled profiles; only 21 InternVL2 conditions cover all positions. These summaries are unavailable for the remaining images in the main cohort and are handled through training-fold mean imputation and missingness indicators. Both predictors use identical feature coverage. The new cohort also has 12 images designated in advance for this measurement. Results for the subcohort with complete summary coverage are reported separately.

The full-visual-removal trajectory in layer-aware modeling consists of joint-preservation outcomes at seven layers. It determines the relative position of the first successful layer, the distance from the current layer to that position, and a right-censoring indicator. These features use only full-removal observations, without accessing sufficiency probabilities or controlled-selector labels, and summarize depth-dependent changes in the effects of visual removal.

\subsection{Learners and data splits}
Fixed logistic regression uses an L2 penalty of 1.0 with an unpenalized intercept. Sorted image IDs are assigned to eight folds in round-robin order, and all preprocessing parameters are estimated on the training folds.
Augmented logistic regression adds categorical encoding for seven layers, layer-by-selector interactions, and interactions of layer with full-removal NLL, KL, dense NLL, and query-similarity top-$K$ mass. The predictor with sufficiency probability also includes its interaction with layer. Candidate L2 penalties are 100, 10, 1, and 0.1.
Constrained gradient-boosted trees use the histogram gradient boosting classifier in scikit-learn~\cite{pedregosa2011sklearn}, with a learning rate of 0.05, a minimum of 64 samples per leaf, and at most 8 leaf nodes. The search covers eight combinations of depth (2 or 3), iterations (100 or 200), and L2 penalty (10 or 1).

Layer-aware modeling uses image-grouped splits with eight outer and four inner folds. Inner-fold log loss for the baseline without sufficiency probability, averaged equally over images, selects the learner family. The two predictors are then tuned separately within the same search space. Direct evaluation on new images uses the final predictors trained on all 32 original images; refitting within the new cohort assesses the stability of incremental information under retraining. Predictors are fitted separately for each model. Image pairing permits comparison on the same inputs, while parameters remain model-specific.

\subsection{Losses and statistical units}
Let $\mathcal A_x$ be the set of evaluated pruning actions for image $x$, and let $\widehat z_a$ be the predicted failure probability. The image-level loss is
\begin{equation}
\begin{split}
\mathcal L_x=-\frac1{|\mathcal A_x|}\sum_{a\in\mathcal A_x}
[&z_a\log\widehat z_a\\
&+(1-z_a)\log(1-\widehat z_a)].
\end{split}
\end{equation}
Overall loss averages equally across images. Brier score likewise averages $(\widehat z_a-z_a)^2$ within each image and then equally across images. Bootstrap differences preserve pairing between predictors and resample images. All bootstrap intervals are conditional on existing out-of-fold or fixed predictions, without retraining within each resample.

\section{Supplementary Prediction Results}
\label{app:predictionresults}
\subsection{Feature ablation and paired losses}
Table~\ref{tab:linear} reports out-of-fold losses as features are added, and Table~\ref{tab:linearci} compares increments, intervals, and directional support for the two prediction settings. For Qwen2.5-VL, adding query summaries raises log loss from $M_1$ to $M_2$; adding sufficiency probability in $M_3$ lowers it below both.

Under fixed logistic regression, log loss decreases from 0.4786, 0.4371, and 0.5554 to 0.4454, 0.3796, and 0.4519 for Qwen2.5-VL, InternVL2, and LLaVA, respectively, corresponding to reductions of 6.9\%, 13.2\%, and 18.6\%. Together with the main-text layer-aware results, log loss and Brier point estimates decrease in both settings, with relative log loss reductions of 4.0\%--18.6\%. For InternVL2 and LLaVA, intervals for both loss differences lie below zero in both settings. Some Qwen2.5-VL intervals span zero; gains in the original cohort concentrate in early and middle layers, with reversed point estimates in late layers. Layer-aware modeling lowers baseline loss for all three models, and adding sufficiency probability lowers it further. The prespecified alternative fold assignment retains negative differences.

\begin{table}[!htbp]
\centering\normalsize
\caption{Feature ablation with fixed logistic regression. All feature sets use the same action records and eight image-grouped folds.}
\label{tab:linear}
\begin{tabular}{llcc}
\toprule
Model & Features & Log loss & Brier\\
\midrule
Qwen2.5-VL & $M_0$ & 0.50161 & 0.16507\\
 & $M_1$ & 0.47349 & 0.15330\\
 & $M_2$ & 0.47855 & 0.15499\\
 & $M_3$ & 0.44545 & 0.14388\\
InternVL2 & $M_0$ & 0.47407 & 0.15377\\
 & $M_1$ & 0.45001 & 0.14825\\
 & $M_2$ & 0.43711 & 0.14158\\
 & $M_3$ & 0.37962 & 0.12080\\
LLaVA & $M_0$ & 0.55845 & 0.19022\\
 & $M_1$ & 0.56984 & 0.17698\\
 & $M_2$ & 0.55543 & 0.17321\\
 & $M_3$ & 0.45187 & 0.13665\\
\bottomrule
\end{tabular}
\end{table}

\begin{table}[!htbp]
\centering\normalsize
\caption{Paired loss differences after adding sufficiency probability, with 95\% intervals. Fixed logistic regression compares $M_3-M_2$. The final column gives directional support for improvements in log loss and Brier, respectively.}
\label{tab:linearci}
\setlength{\tabcolsep}{2pt}
\begin{tabular}{lccc}
\toprule
Setting & \makecell{$\Delta$ log loss\\{}[95\% CI]} & \makecell{$\Delta$ Brier\\{}[95\% CI]} & \makecell{Directional\\support}\\
\midrule
\multicolumn{4}{l}{Qwen2.5-VL}\\
Fixed logistic & \makecell{$-0.03311$\\$[-0.06128,-0.00175]$} & \makecell{$-0.01112$\\$[-0.02152,0.00079]$} & \makecell{98.16\% /\\96.82\%}\\
Layer-aware & \makecell{$-0.0172$\\$[-0.0348,0.0017]$} & \makecell{$-0.0058$\\$[-0.0123,0.0010]$} & \makecell{96.26\% /\\95.00\%}\\
\multicolumn{4}{l}{InternVL2}\\
Fixed logistic & \makecell{$-0.05749$\\$[-0.08698,-0.03018]$} & \makecell{$-0.02077$\\$[-0.03207,-0.01049]$} & \makecell{100\% /\\100\%}\\
Layer-aware & \makecell{$-0.0672$\\$[-0.1012,-0.0398]$} & \makecell{$-0.0211$\\$[-0.0327,-0.0113]$} & \makecell{100\% /\\100\%}\\
\multicolumn{4}{l}{LLaVA}\\
Fixed logistic & \makecell{$-0.10356$\\$[-0.13709,-0.07258]$} & \makecell{$-0.03656$\\$[-0.04779,-0.02640]$} & \makecell{100\% /\\100\%}\\
Layer-aware & \makecell{$-0.0797$\\$[-0.1069,-0.0537]$} & \makecell{$-0.0310$\\$[-0.0408,-0.0216]$} & \makecell{100\% /\\100\%}\\
\bottomrule
\end{tabular}
\end{table}

\begin{table}[!htbp]
\centering\normalsize
\caption{Supplementary refitting results within the new GQA cohort. This uses the same 32 images as the direct new-image evaluation in the main text, addressing prediction under a different training setting.}
\label{tab:refit}
\begin{tabular}{lcccc}
\toprule
Model & \makecell{Without\\$\psuf$} & \makecell{With\\$\psuf$} & \makecell{$\Delta$ log loss\\{}[95\% CI]} & \makecell{Relative\\reduction}\\
\midrule
Qwen2.5-VL & 0.3906 & 0.3379 & \makecell{$-0.0527$\\$[-0.0662,-0.0390]$} & 13.5\%\\
InternVL2 & 0.3922 & 0.3507 & \makecell{$-0.0416$\\$[-0.0666,-0.0190]$} & 10.6\%\\
LLaVA & 0.4604 & 0.4035 & \makecell{$-0.0569$\\$[-0.0753,-0.0385]$} & 12.4\%\\
\bottomrule
\end{tabular}
\end{table}

\begin{table}[!htbp]
\centering\normalsize
\caption{Brier differences on the new cohort with 95\% intervals. Negative values indicate improvement after adding sufficiency probability.}
\label{tab:brier}
\begin{tabular}{lcc}
\toprule
Model & \makecell{Direct evaluation of\\fixed predictors} & \makecell{Nested refitting within\\the new cohort}\\
\midrule
Qwen2.5-VL & $-0.0143\;[-0.0191,-0.0100]$ & $-0.0199\;[-0.0257,-0.0138]$\\
InternVL2 & $-0.0159\;[-0.0223,-0.0096]$ & $-0.0137\;[-0.0222,-0.0055]$\\
LLaVA & $-0.0253\;[-0.0323,-0.0186]$ & $-0.0221\;[-0.0286,-0.0154]$\\
\bottomrule
\end{tabular}
\end{table}

\subsection{Complete-summary subcohort and new-cohort refitting}
In the preselected 12-image subcohort with complete summary coverage, layer-aware log loss differences are $-0.0310\;[-0.0691,0.0176]$ for Qwen2.5-VL, $-0.0484\;[-0.0805,-0.0157]$ for InternVL2, and $-0.0875\;[-0.1490,-0.0365]$ for LLaVA. Qwen2.5-VL has 91.0\% directional support for improvement. This subcohort assesses the effect of single-token summary coverage; each condition still estimates its summaries from at most 256 positions.

Table~\ref{tab:refit} reports refitting within the new cohort. The predictor without sufficiency probability selects gradient-boosted trees in every outer fold. Log loss improves for 29, 26, and 25 images in the three models, respectively. Table~\ref{tab:brier} reports Brier differences for both fixed prediction and refitting on the new cohort. All intervals are negative, with 100\% directional support in both analyses.

\section{Supplementary Evaluation of Full Pruning Methods}
\label{app:fullmethods}
\subsection{Native pipelines and predictor configurations}
FastV retains 144/576 visual positions after decoder block 9 of base LLaVA, following the authors' attention-ranking pipeline. VisionZip uses the authors' LLaVA-1.5 inference configuration with dominant/contextual counts of 54/10, yielding 64 visual tokens. SparseVLM uses the authors' v1.5 branch and the three-stage configuration \texttt{RETAIN\_TOKN=128}, ultimately retaining 39 visual tokens. Each method is compared with a dense counterpart using identical model weights. Joint-preservation success $Y=1$ requires the generated answer, task score, reference-answer NLL increase, and dense-to-pruned output KL to satisfy Eq.~\eqref{eq:success}; the failure label is $1-Y$. Table~\ref{tab:fullmethods} compares failure prediction within each method, retaining its native intervention positions and final budget.

For each full pruning method, we fit two L2-regularized logistic regression predictors, B0 and B1, to joint-preservation failure. The L2 penalty is fixed at 1.0, and the intercept is unpenalized. B0 uses question-type encoding and 13 continuous or binary summaries: question length in characters; dense output NLL, entropy, and the top-two probability gap; task preservation, NLL increase, KL, and joint preservation under full visual removal; and the mean, standard deviation, normalized entropy, top-$K$ mass, and Gini coefficient of query similarity. B1 adds smoothed sufficiency probability without further transformation. Both predictors use the same samples, regularization penalty, and image splits, with separately fitted coefficients.

\subsection{Metric tests and question-type analysis}
Table~\ref{tab:transplantmetrics} reports Brier score, AP, and relative loss changes. Nominal two-sided $p$ values for architecture-adapted configurations use $t$ tests on image-mean loss differences. Main-text intervals use image-cluster bootstrap resampling of fixed predictions. Average precision (AP) groups tied scores at common thresholds.

Question-type analyses compute paired losses for attribute, relation, and object questions. Two-sided tests for both losses use image-mean differences, and intervals use 5,000 image-cluster bootstrap resamples.

\subsection{Components of joint-preservation failure}
\label{app:methoddiagnostics}
Table~\ref{tab:failurecomponents} decomposes joint-preservation failures in the first cohort. For the three methods, 19, 26, and 24 questions satisfy task preservation but fail because NLL or KL exceeds its threshold. The joint label therefore captures both answer and distribution preservation, while task correctness is evaluated against reference answers.

\begin{table}[!htbp]
\centering\normalsize
\caption{Components of joint-preservation failure in the first full-method cohort, with 96 questions per method. Task-preservation failure means disagreement with the normalized dense answer or a lower task score. NLL and KL exceedances mean an increase above 0.10 and divergence above 0.05 nats/token, respectively; components can overlap. The final column counts task-preserving cases that fail joint preservation.}
\label{tab:failurecomponents}
\setlength{\tabcolsep}{5pt}
\begin{tabular}{lccccc}
\toprule
Method & \makecell{Joint\\failure} & \makecell{Task\\failure} & \makecell{NLL\\exceeded} & \makecell{KL\\exceeded} & \makecell{Task pass,\\joint fail}\\
\midrule
FastV & 41 & 22 & 24 & 13 & 19\\
VisionZip & 50 & 24 & 12 & 45 & 26\\
SparseVLM & 40 & 16 & 14 & 31 & 24\\
\bottomrule
\end{tabular}
\end{table}

\subsection{Method-level cross-image risk alerts}
Table~\ref{tab:fullalert} gives method-level alerts for the LLaVA transfer cohort. Each method flags the 32 of 96 questions with the lowest sufficiency probabilities. RD is the failure rate in this group minus that in the remaining 64 questions. Resampling uses common image clusters across methods and reranks the existing scores. All three point estimates have the same direction. The SparseVLM interval is above zero, while the FastV and VisionZip intervals span zero. The equally weighted risk difference is $0.1719\;[0.0365,0.3021]$, converted to percentage points in Table~\ref{tab:transplantrisk}.

\par\addvspace{\intextsep}
\noindent\begin{minipage}{\linewidth}
\singlespacing\centering\normalsize
\captionof{table}{Method-level risk alerts on the LLaVA transfer cohort. Each method flags 32 questions. RD is the failure rate for flagged questions minus that for the remainder; hits count failures among flagged questions.}
\label{tab:fullalert}
\setlength{\tabcolsep}{3pt}
\begin{tabular}{lccc}
\toprule
Method & Hits & RD & 95\% CI\\
\midrule
FastV & 14/32 & $+0.1094$ & $[-0.0473,+0.3281]$\\
VisionZip & 20/32 & $+0.0625$ & $[-0.1719,+0.2656]$\\
SparseVLM & 22/32 & $+0.3438$ & $[+0.1562,+0.5156]$\\
\midrule
Equal-weight mean & -- & $+0.1719$ & $[+0.0365,+0.3021]$\\
\bottomrule
\end{tabular}
\end{minipage}
\par\addvspace{\intextsep}

Risk localization for Qwen2.5-VL and InternVL2 uses complete cohorts of 32 images with three questions each, giving 96 questions per model, method, and cohort. The main and transfer cohorts contain 64 distinct images and 192 questions in total. Reusing these questions across two models and three methods yields 1,152 evaluation records, with images remaining the resampling unit. All four macro risk differences for these two models in Table~\ref{tab:transplantrisk} have 100.00\% directional support: every resampled risk difference is positive.

\section{Successful-Set Structure and Measurement Scope}
\subsection{Proof of successful-subset disagreement across queries}
\label{app:successdisagreement}
Using the common subset space $\Omega$ and uniform distribution $U$ from the main text, write the two success labels as $Y_1(S)$ and $Y_2(S)$. Since each label is 0 or 1, $|Y_1(S)-Y_2(S)|$ is exactly the indicator of disagreement between preservation outcomes. Hence,
\begin{equation}
\begin{split}
|p_{q_1}-p_{q_2}|
&=\left|\mathbb E_U[Y_1(S)-Y_2(S)]\right|\\
&\leq\mathbb E_U[|Y_1(S)-Y_2(S)|]\\
&=U(A_{q_1}\triangle A_{q_2}),
\end{split}
\end{equation}
which proves Proposition~\ref{prop:successdisagreement}. Equality holds when one successful set contains the other. Conversely, equally sized disjoint successful sets have the same probability, showing that the full overlap structure requires joint-response information.

This relation concerns population probabilities $p_q$. Substituting finite-sample estimates $\widehat p_q$ yields an estimate of the lower bound that remains subject to sampling error. The main text compares query-specific curves against independent replication error.

\subsection{Minimum concentration of successful selection distributions}
\label{app:selectionconcentration}
Fix a question and its intervention conditions, with $p_q>0$. Let $Q$ be any subset selection distribution on $\Omega$ satisfying $Q(A_q)=1$, so that every selected subset meets the joint-preservation criterion. The uniform distribution on the successful set is $U_{A_q}(S)=U(S)/p_q$ for $S\in A_q$. By the definition of KL divergence,
\begin{equation}
\begin{split}
D_{\mathrm{KL}}(Q\Vert U)
&=\sum_{S\in A_q}Q(S)\log\frac{Q(S)}{p_qU_{A_q}(S)}\\
&=D_{\mathrm{KL}}(Q\Vert U_{A_q})-\log p_q.
\end{split}
\label{eq:selectionkldecomposition}
\end{equation}
The first term is nonnegative and vanishes when $Q=U_{A_q}$, giving
\begin{equation}
\min_{Q:\,Q(A_q)=1}D_{\mathrm{KL}}(Q\Vert U)
=-\log p_q.
\label{eq:selectionconcentration}
\end{equation}
Logarithms are natural, and divergence is measured in nats. When $p_q=1$, uniform selection satisfies the constraint and the minimum divergence is zero. When $p_q=0$, the successful set is empty and no selection distribution satisfies the constraint.

This identity translates successful-combination abundance into the minimum distributional concentration required for successful selection. A smaller successful region requires a greater departure from uniformity to place all selection mass within it. The KL divergence here is defined over visual-subset selection distributions; the KL in Eq.~\eqref{eq:success} is defined over model output distributions. The minimum divergence constrains selection distributions, while computational cost depends on the selector implementation and available information.

\subsection{Singleton observations do not determine success probability mass}
\label{app:identifiability}
Let $F(S)$ denote the functional response after retaining subset $S$. Without restrictions on the set function, for $2\leq K<N$ one can construct two response functions that agree on the empty set, every singleton, and the full set, but satisfy and fail the success criterion, respectively, on every $K$-element subset. Their endpoint and singleton observations are identical, yet their sufficiency probabilities are 1 and 0. Success mass over fixed-size combinations therefore requires additional response observations or structural assumptions. The prediction experiments in the main text test the information provided by this statistical object beyond visual-information and query-representation summaries.

\subsection{Interpreting successful-set membership statistics}
\label{app:membership}
Under uniform fixed-size sampling, any token $t$ is retained with probability $\rho_{\mathrm{eff}}=K/N$. If the success probability satisfies $\psuf>0$, then
\begin{equation}
\begin{split}
\mathbb P(t\in S\mid Y=1)
&=\frac{\mathbb P(t\in S,Y=1)}{\psuf}\\
&\leq\min\left\{1,\frac{\rho_{\mathrm{eff}}}{\psuf}\right\}.
\end{split}
\label{eq:membershipbound}
\end{equation}
When the effective retention ratio is approximately 0.25 and the true sufficiency probability is at least 0.80, any token appears in successful subsets with probability at most approximately 0.3125. Requiring this probability to reach 0.95 implies $\psuf\leq\rho_{\mathrm{eff}}/0.95$, approximately 0.263. High success mass thus limits the presence of nearly mandatory individual tokens.

The original membership analysis defines stable-core candidates through high occurrence frequency and enrichment checks in both sampling replicates, using the first condition meeting the high-sufficiency screening criterion for each input. Table~\ref{tab:membership} reports membership statistics for both tasks. None of the three models yields a core token satisfying the 0.95 frequency threshold in both replicates, consistent with the constraint on high-sufficiency regions in Eq.~\eqref{eq:membershipbound}. The bound concerns population probabilities; observed membership frequencies also depend on finite sampling and rounding $K$ upward.

\begin{table}[!htbp]
\centering\normalsize
\caption{Membership diagnostics for high-sufficiency inputs. $\Delta_J$ is the deviation of mean pairwise Jaccard overlap among successful subsets from the conditional uniform combinatorial baseline. Pooled means weight tasks by their numbers of analyzable inputs.}
\label{tab:membership}
\setlength{\tabcolsep}{3pt}
\begin{tabular}{lcccc}
\toprule
Model & GQA & TextVQA & Total & Mean $\Delta_J$\\
\midrule
Qwen2.5-VL & 28 & 6 & 34 & $-6.73\!\times\!10^{-6}$\\
InternVL2 & 26 & 26 & 52 & $-2.29\!\times\!10^{-5}$\\
LLaVA & 31 & 30 & 61 & $3.42\!\times\!10^{-5}$\\
\bottomrule
\end{tabular}
\end{table}

Mean overlap deviations are close to the conditional combinatorial baseline. These statistics describe membership and pairwise overlap in the successful set; identifying higher-order combinatorial structure requires direct analysis of joint responses to multiple tokens.

\section{Cohort Construction and Study Design}
\label{app:provenance}
\subsection{Task selection}
\begin{sloppypar}
Experiments focus on GQA and TextVQA. Preliminary work also considered RefCOCOg~\cite{mao2016refcocog}. One input in the Qwen2.5-VL run produced an invalid dense output, and none of the six LLaVA interface-check inputs produced parseable coordinates. Localization measurements were subsequently discontinued.
\end{sloppypar}

\subsection{Original cohort and new images}
Sampling of the original cohort did not exclude all images used in historical pilot experiments. The Qwen2.5-VL budget supplement and original-cohort prediction study were follow-up analyses conducted after inspecting some main-experiment results. The two prediction analyses use the same images and selector labels, assessing incremental sufficiency information through different feature representations and fitting procedures.

Images, questions, and analyses for the new cohort were specified before measurement. Final predictors for the original cohort were trained on September 24, 2026 and applied directly to the new cohort later that day. Nested refitting within the same new cohort was performed the following day. Direct evaluation tests the original predictive relationships on new images; refitting evaluates newly fitted predictors on that same image set.

Direct comparison of image IDs excluded 725 previously used IDs recorded in 73 readable historical sources, together with the original training images for all three models. This comparison cannot cover historical exploration absent from those records. Each model completed 96 question--answer pairs, 672 conditions, and 2,016 main-selector records, with no measurement failures or replacement images. The 12 images for single-token summaries were selected before measurement.
\end{document}